\documentclass[letterpaper]{article} % DO NOT CHANGE THIS
\usepackage[preprint]{aaai2027}
\usepackage[hyphens]{url}  % DO NOT CHANGE THIS
\usepackage{graphicx} % DO NOT CHANGE THIS
\usepackage{natbib}  % DO NOT CHANGE THIS AND DO NOT ADD ANY OPTIONS TO IT
\usepackage{caption} % DO NOT CHANGE THIS AND DO NOT ADD ANY OPTIONS TO IT
\usepackage{booktabs}
\usepackage{amsmath,amssymb,amsfonts,bm}
\usepackage{array}
\usepackage{tabularx}

\graphicspath{{figures/}}

\newcommand{\R}{\mathbb{R}}
\newcommand{\Dobs}{\mathcal{D}_{\mathrm{obs}}}
\newcommand{\Sset}{\mathcal{S}}
\newcommand{\Cset}{\mathcal{C}}

\newcommand{\Gset}{\mathcal{G}}
\newcommand{\norm}[1]{\lVert #1 \rVert}

\newcolumntype{P}[1]{>{\raggedright\arraybackslash}p{#1}}
\newcolumntype{C}[1]{>{\centering\arraybackslash}p{#1}}
\newcolumntype{Y}{>{\raggedright\arraybackslash}X}
\newcounter{algorithm}
\title{Freeze, Then Select: Structured Field Adapters and Stability-Validated Weak Selection for PDE Discovery from Sparse Observations}
\author{
    Juncheng Zhong\textsuperscript{\rm 1},
    Chenghuang Shen\textsuperscript{\rm 3,\rm 4},
    Jianfeng Liu\textsuperscript{\rm 4},
    Zhengdong Xiao\textsuperscript{\rm 4},\\
    Longjiu Luo\textsuperscript{\rm 4},
    Qianrong Wang\textsuperscript{\rm 4},
    Wenjun Xu\textsuperscript{\rm 5,\rm 6,*},
    Wenlian Lu\textsuperscript{\rm 1,\rm 2,\rm 3,*}
}
\affiliations{
    \textsuperscript{\rm 1}School of Mathematical Sciences, Fudan University, Shanghai 200433, China\\
    \textsuperscript{\rm 2}Center for Applied Mathematics, Fudan University, Shanghai 200438, China\\
    \textsuperscript{\rm 3}Shanghai Center for Mathematical Sciences, Fudan University, Shanghai 200438, China\\
    \textsuperscript{\rm 4}Alibaba Group, Hangzhou, China\\
    \textsuperscript{\rm 5}Shanghai Institute of Technical Physics, CAS, Shanghai 200083, China\\
    \textsuperscript{\rm 6}National Key Laboratory of Infrared Detection Technologies, Shanghai Institute of Technical Physics, CAS, Shanghai 200083, China\\
    \textsuperscript{*}Corresponding authors:
    wenlian@fudan.edu.cn, xuwenjun@mail.sitp.ac.cn
}

\begin{document}

\maketitle

\begin{abstract}
PDE discovery from sparse observations requires reconstructing a continuous field and selecting the correct differential terms. Our analysis of optimization paths in coupled neural PDE discovery reveals three behaviors: the exact support can persist to the end of training, appear only transiently, or fail to emerge. To decouple equation selection from neural optimization, we develop a freeze-then-select method combining a structured field adapter with Stability-Validated Weak Selection (SVWS). Trained from observations without a PDE residual, the adapter factorizes the field into learned spatial features and temporal coefficients represented by cubic splines. After freezing the field, SVWS identifies recurrent terms across independent weak-form systems, refits candidate supports, and selects the final equation on held-out weak-form systems. Beyond fixed libraries, we apply the same principle to expressions generated by genetic programming and recover the power-law form of an unknown nonlinear diffusion function from sparse, noisy observations. Across all six sparse MDBench regimes, our method attains the highest exact support recovery rate, with its clearest gains over classical and neural baselines on challenging Kuramoto--Sivashinsky dynamics.
\end{abstract}

\section{Introduction}

Data-driven discovery of governing equations seeks parsimonious dynamical laws from measurements of physical systems \citep{bongard2007automated,schmidt2009distilling,brunton2016discovering}. For PDEs, sparse-regression methods such as PDE-FIND select active differential terms from prescribed libraries \citep{rudy2017data,schaeffer2017learning}, while weak formulations reduce sensitivity to pointwise differentiation \citep{schaeffer2017sparseintegral,messenger2021weak}. These approaches require field values, derivatives, or weak integrals on a suitable space--time representation. When observations cover only a subset of spatial sensors or time frames, those quantities are not directly available. PDE discovery from sparse observations therefore entails two linked tasks: reconstructing a continuous field and selecting the governing terms from the quantities induced by that reconstruction.

\begin{figure}[t]
  \centering
  \includegraphics[width=\columnwidth]{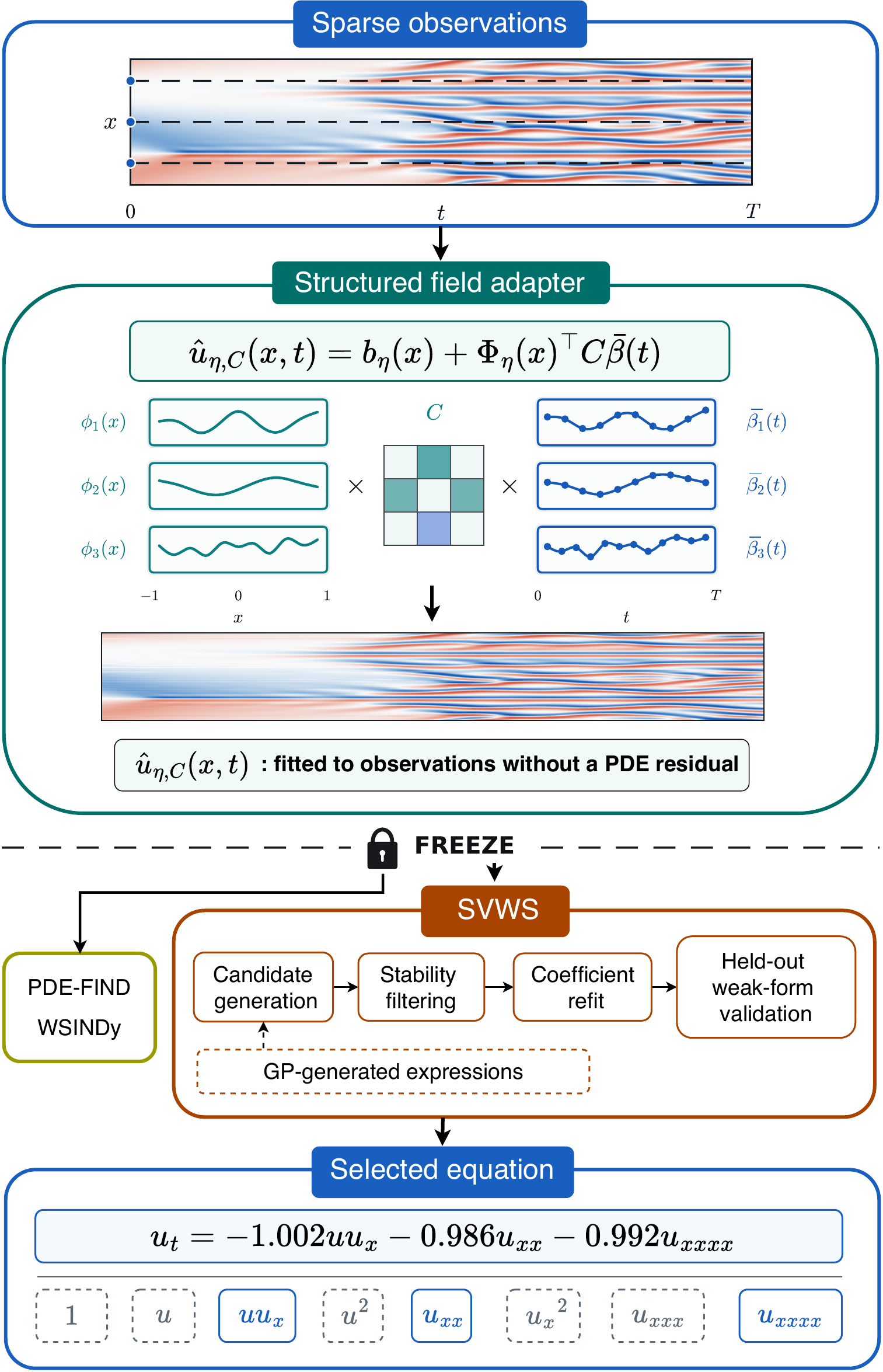}
  \caption{Freeze, then select. A structured adapter fits sparse observations without a PDE residual and is frozen before equation discovery. SVWS selects both fixed-library supports and GP-generated expressions; the frozen field can also be passed to classical selectors.}
  \label{fig:overview}
\end{figure}

Neural surrogates address reconstruction by providing continuous, differentiable fields from limited observations \citep{both2021deepmod,chen2021physics,stephany2024pdelearn,stephany2024weak}. Many neural PDE-discovery methods couple the surrogate to equation coefficients or operator selection. This coupling can be effective: the evolving equation can regularize reconstruction, while the surrogate supplies derivatives or weak integrals away from sensors. The reconstructed field and inferred support, however, evolve together along a single optimization path. Our analysis of these paths reveals three behaviors in coupled neural PDE discovery (Fig.~\ref{fig:joint_mechanism}): the exact support can persist to the end of training, appear only transiently, or fail to emerge.

These observations motivate our freeze-then-select design, which separates equation selection from neural surrogate training (Fig.~\ref{fig:overview}). We fit a continuous field from observations without a PDE residual, freeze it, and evaluate candidate equations on the same reconstruction. Earlier two-stage methods demonstrate the value of this separation: DL-PDE applies sparse regression after fitting a neural field, while an integral-form method performs genetic search on quantities obtained from a pretrained surrogate \citep{xu2021dlpde,xu2021integralpde}. For sparse observations, however, two questions remain: how to reconstruct a field that yields informative differential or weak-form quantities, and how to reduce sensitivity to the choice of weak-form system and sparsity threshold. We address these questions with a structured field adapter and Stability-Validated Weak Selection (SVWS).

\begin{figure*}[!t]
  \centering
  \includegraphics[width=0.96\textwidth]{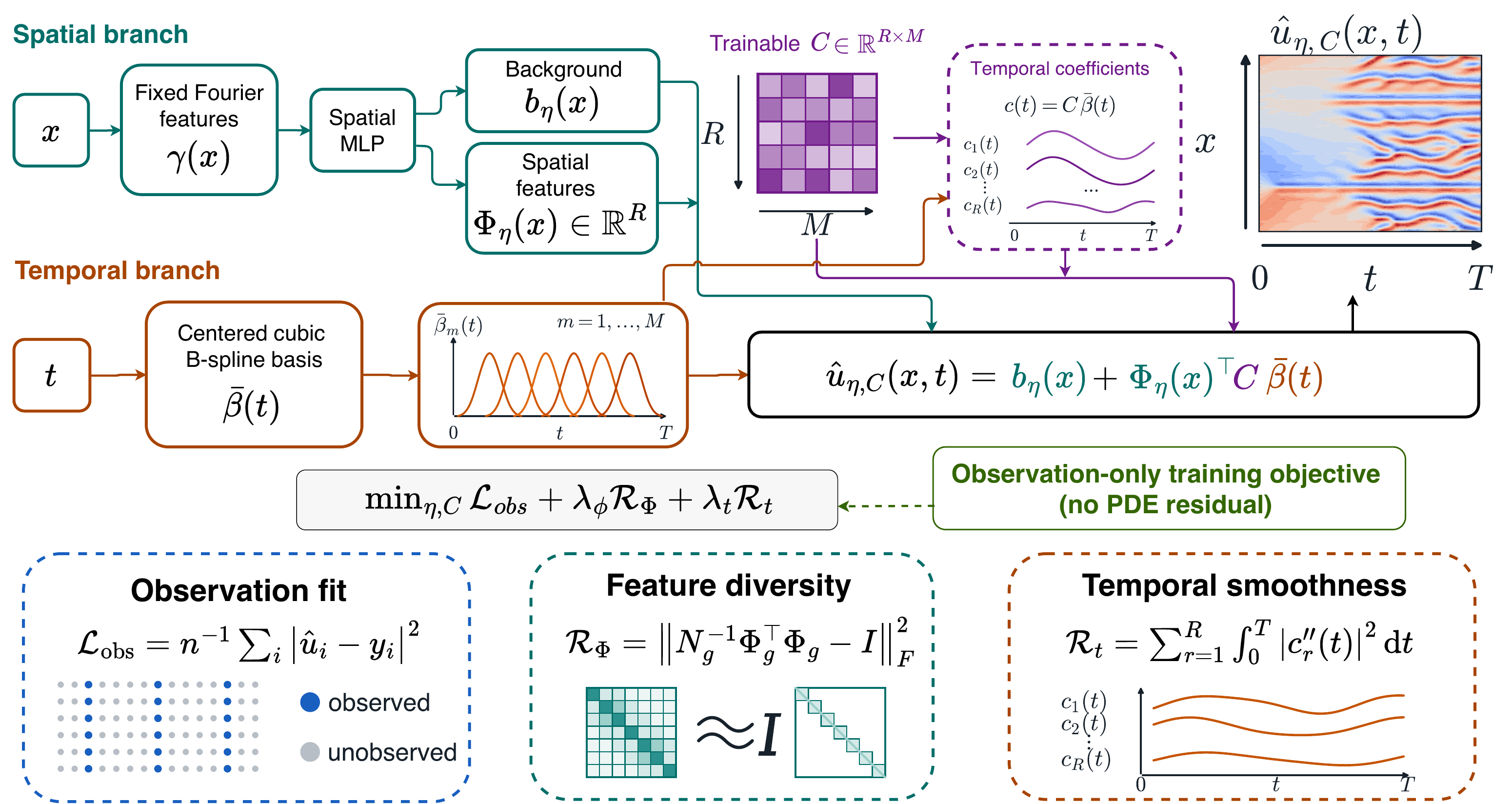}
  \caption{Structured field adapter. A spatial network produces a background and spatial features with temporal coefficients represented in a centered cubic B-spline basis. The adapter is fit without a PDE residual and frozen before equation selection.}
  \label{fig:adapter_architecture}
\end{figure*}

The structured adapter factorizes the field into learned spatial features and temporal coefficients represented in a cubic B-spline basis (Fig.~\ref{fig:adapter_architecture}). This design follows the familiar view of coherent dynamics as spatial patterns with amplitudes that evolve over time \citep{sirovich1987turbulence,holmes1996turbulence}. The learned features capture complex spatial structure, while the spline representation provides smooth temporal evolution and analytic time derivatives \citep{sun2021pisl,sun2022bayesianspline}. After freezing the field, SVWS generates candidate supports across independently constructed weak-form systems and sparsity thresholds, retains recurrent terms, refits each support, and selects the final equation on held-out weak-form systems. In this selection stage, integration by parts avoids high-order pointwise derivatives of the reconstruction \citep{messenger2021weak,tang2023weakident}, while shifted test-function grids provide multiple views of the same frozen field for stability assessment and held-out validation. Beyond fixed libraries, an extension of SVWS refits and validates symbolic expressions generated by genetic programming (GP) on independent weak-form systems from the frozen field.

We evaluate the method on sparse MDBench regimes spanning KdV, Kuramoto--Sivashinsky (KS), and two-dimensional advection--diffusion, with observations restricted to subsets of fixed spatial sensors or time frames \citep{bideh2026mdbench}. Our method attains the highest rate of exact support recovery in each of the six regimes, with its clearest gains over classical and neural baselines on KS (Table~\ref{tab:fixed_library_main}). Adapter ablations and selector sensitivity analyses examine the two stages of the method (Table~\ref{tab:structure_ablation_main}). Beyond fixed libraries, the SVWS extension recovers the power-law form of an unknown nonlinear diffusion function from sparse, noisy observations (Table~\ref{tab:openform_capability}).

\noindent\textbf{Contributions.}
(i) We develop a freeze-then-select PDE-discovery method whose structured field adapter couples learned spatial features to explicit temporal spline coefficients and is trained from observations without a PDE residual.
(ii) We introduce SVWS, which separates support generation, coefficient refitting, and validation across independently constructed weak-form systems.
(iii) We analyze optimization paths in coupled neural PDE discovery, showing that the exact support can persist to the end of training, appear only transiently, or fail to emerge.
(iv) We extend SVWS beyond fixed libraries to expressions generated by GP, demonstrating recovery of the power-law form of an unknown nonlinear diffusion function from sparse, noisy observations.

\section{Related Work}

\noindent\textbf{Sparse regression and model selection.}
SINDy identifies parsimonious governing equations from predefined libraries \citep{brunton2016discovering}, and PDE-FIND extends sparse identification to PDEs \citep{rudy2017data}. Later work develops implicit and relaxed formulations, ensemble methods, information criteria, and approaches to noisy or incomplete data \citep{mangan2017model,zheng2019sr3,kaheman2020sindypi,reinbold2020using,fasel2022ensemble,xu2023pic}. Stability selection retains terms that recur across resampled fits, while the one-standard-error rule selects the simplest model within one standard error of the minimum validation error \citep{meinshausen2010stability,hastie2009elements}. SVWS adapts these ideas to supports generated across independently constructed weak-form systems.

\noindent\textbf{Weak and integral formulations.}
Weak-form PDE discovery transfers derivatives from the field to smooth test functions, reducing reliance on pointwise differentiation. Variational system identification, integral sparse selection, WSINDy, and WeakIdent use this principle to improve robustness to noise \citep{wang2019vsi,schaeffer2017sparseintegral,messenger2021weak,tang2023weakident}. We apply weak-form selection after freezing the reconstructed field.

\noindent\textbf{Neural surrogates for PDE discovery.}
Physics-informed neural networks (PINNs) fit differentiable surrogates with data and physics objectives \citep{raissi2019physics,karniadakis2021physics}; balancing these objectives remains an important optimization challenge \citep{wang2021gradient,krishnapriyan2021failure}. Coupled neural discovery methods, including PDE-Net, DeepMoD, PINN-SR, PDE-LEARN, and Weak-PDE-LEARN, learn trainable representations together with differential operators or coefficients \citep{long2018pdenet,long2019pdenet,both2021deepmod,chen2021physics,stephany2024pdelearn,stephany2024weak}. Two-stage alternatives separate reconstruction from discovery: DL-PDE applies sparse regression to a fitted neural field, while an integral-form method performs genetic search on quantities obtained from a pretrained surrogate \citep{xu2021dlpde,xu2021integralpde}. We build on this separation with a structured adapter for reconstruction and SVWS after freezing.

\noindent\textbf{Structured field representations.}
Reduced-order models describe coherent dynamics through spatial modes and time-dependent coefficients \citep{sirovich1987turbulence,holmes1996turbulence}. Coordinate networks provide continuous fields, with Fourier features and periodic activations improving high-frequency representation \citep{tancik2020fourier,sitzmann2020siren}. Spline-based methods combine smooth reconstruction with sparse equation discovery \citep{sun2021pisl,sun2022bayesianspline}. Our adapter follows this structure, learning spatial features with temporal coefficients represented in a spline basis.

\noindent\textbf{Symbolic PDE discovery.}
Symbolic regression searches compositional expressions beyond fixed candidate libraries \citep{koza1992genetic,bongard2007automated,schmidt2009distilling,cranmer2023interpretable}. PDE discovery beyond fixed libraries has used surrogate-assisted and symbolic genetic algorithms \citep{xu2020dlgapde,zhang2022sgapde}, physics-informed genetic programming \citep{cohen2024physicsgp}, and closed-form search \citep{kacprzyk2023dcipher}. Recent approaches also use reinforcement learning, abductive learning, differentiable weak-form networks, and generative models \citep{du2024discover,gao2025ablpde,li2026weakpdenet,xu2025eqgpt}. In our setting, genetic programming proposes expressions for an unknown nonlinear diffusion function, and the SVWS extension selects among them using weak-form systems from the frozen field.

\section{Freeze-Then-Select PDE Discovery}

Our freeze-then-select method fits a continuous field to sparse observations without a PDE residual and then freezes it, so every candidate equation is evaluated on the same reconstruction. The structured field adapter provides this reconstruction, and Stability-Validated Weak Selection (SVWS) assigns support generation, coefficient refitting, and validation to independent weak-form systems. The frozen field can also be passed to classical PDE selectors. Beyond fixed libraries, the same refitting and validation principle applies to expressions generated by genetic programming.

\noindent\textbf{Problem setting.}
Let $u:\Omega\times[0,T]\to\R$ be an unknown scalar field on $\Omega\subset\R^d$, observed through sparse, possibly noisy samples
\begin{equation}
  \Dobs = \{(x_i,t_i,y_i)\}_{i=1}^{n},
  \qquad
  y_i = u(x_i,t_i)+\epsilon_i,
\end{equation}
where $\epsilon_i$ denotes measurement noise. The sampling pattern may retain only a subset of spatial locations or time slices. In the fixed-library setting, the governing equation is assumed to be sparse in a prescribed library of $J$ candidate terms:
\begin{equation}
  \begin{aligned}
  \partial_t u
    &= \sum_{j=1}^{J}\xi_j^\star
       \Theta_j(u,\nabla u,\nabla^2u,\ldots),\\
  \Sset^\star
    &= \{j:\xi_j^\star\neq0\}.
  \end{aligned}
  \label{eq:pde}
\end{equation}
Given $\Dobs$, the goal is to recover $\Sset^\star$ and its coefficients; support recovery is exact when $\widehat{\Sset}=\Sset^\star$. Sparse samples do not directly provide the continuous field, derivatives, or weak integrals required for equation selection. We therefore reconstruct a continuous \emph{field adapter} $\hat u$ from which these quantities can be evaluated.

\noindent\textbf{Structured field adapter.}
The adapter represents temporal evolution through smooth coefficients multiplying learned spatial features. After affine rescaling of $x$, a spatial network receives the fixed Fourier encoding $\gamma(\widetilde{x})=[\sin(B\widetilde{x}),\cos(B\widetilde{x})]$ and outputs a background $b_\eta(x)$ and features $\Phi_\eta(x)\in\R^R$. The rows of $B$ are sampled once from a standard Gaussian distribution and then fixed.

Let $\beta(t)\in\R^M$ be a clamped cubic B-spline basis. We center it over the observed times,
\begin{equation}
  \bar\beta(t)
  =
  \beta(t)
  -
  \frac{1}{|\mathcal{T}_{\mathrm{obs}}|}
  \sum_{\tau\in\mathcal{T}_{\mathrm{obs}}}\beta(\tau),
\end{equation}
so the dynamic component has zero empirical mean over the observed times and $b_\eta(x)$ represents the corresponding mean of the reconstructed field. With $C\in\R^{R\times M}$, the reconstructed field is
\begin{equation}
  \hat u_{\eta,C}(x,t)
  =
  b_\eta(x)+\Phi_\eta(x)^\top C\bar\beta(t).
  \label{eq:adapter}
\end{equation}
Here $c_r(t)=C_{r:}\bar\beta(t)$ is the coefficient of spatial feature $r$, while $R$ and $M$ specify the numbers of spatial features and temporal basis functions.

We fit $(\eta,C)$ from observations by
\begin{equation}
  \min_{\eta,C}\;
  \frac{1}{n}\sum_{i=1}^{n}
  |\hat u_{\eta,C}(x_i,t_i)-y_i|^2
  +\lambda_\Phi\mathcal{R}_\Phi
  +\lambda_t\mathcal{R}_t .
  \label{eq:fit}
\end{equation}
The spatial regularizer $\mathcal{R}_\Phi$ penalizes deviations of the empirical feature Gram matrix from the identity, while $\mathcal{R}_t$ penalizes the integrated squared curvature of the temporal coefficients. The objective contains no PDE residual or PDE coefficients. Regularizer definitions and optimization settings are provided in Supplementary Sec.~A.3. After fitting, we freeze $\hat u_{\eta,C}$ and construct every weak system from this shared reconstruction.

\noindent\textbf{Weak systems from the frozen field.}
Following weak-form PDE discovery \citep{messenger2021weak,tang2023weakident}, we construct local weak systems using compactly supported space--time test functions $\psi_\ell$ on interior patches $P_\ell$. Patch centers follow a regular grid whose phase is sampled independently for each weak system, with one row per center. The polynomial kernel, patch geometry, benchmark-specific settings, and formal error bounds are given in Supplementary Sec.~B.1.

Define $\langle f,g\rangle_\ell=\int_{P_\ell}fg\,dx\,dt$, and write each candidate as $\Theta_j(u)=\mathcal{D}_{\alpha_j}q_j(u)$, with the identity operator covering algebraic terms. Integration by parts gives
\begin{equation}
  b_\ell(\hat u)
  =
  -\langle\hat u,\partial_t\psi_\ell\rangle_\ell,
  \quad
  A_{\ell j}(\hat u)
  =
  \langle q_j(\hat u),\mathcal{D}_{\alpha_j}^{*}\psi_\ell\rangle_\ell.
  \label{eq:weak}
\end{equation}
Here $\mathcal{D}_{\alpha_j}^{*}$ denotes the weak adjoint. Because the test function and its required derivatives vanish at the patch boundary, integration by parts eliminates boundary terms and transfers derivatives from $\hat u$ to the test function. Stacking $m$ patches yields
\begin{equation}
  A(\hat u)\xi\approx b(\hat u),
  \qquad
  A(\hat u)\in\R^{m\times J},
  \quad
  b(\hat u)\in\R^m.
  \label{eq:weak_system}
\end{equation}
If $q_j$ is locally Lipschitz on the reconstructed state range, the error in each weak entry is bounded by the local $L_2$ reconstruction error and the norm of the corresponding test-function derivative. Changing the grid phase changes the sampled local patches while keeping the reconstruction fixed, producing separate weak systems for support generation, coefficient refitting, and validation.

\noindent\textbf{Stability-Validated Weak Selection.}
SVWS assigns weak systems constructed from the same frozen field to three separate roles:
\begin{equation}
\begin{gathered}
  \{(A_{\mathrm{gen}}^{r},b_{\mathrm{gen}}^{r})\}_{r=1}^{R_g},\quad
  (A_{\mathrm{fit}},b_{\mathrm{fit}}),\quad
  \{(A_{\mathrm{val}}^{s},b_{\mathrm{val}}^{s})\}_{s=1}^{R_v}.
\end{gathered}
\label{eq:weak_splits}
\end{equation}
Each system uses a regular patch grid with an independently sampled phase. Generation systems propose supports, the fit system estimates their coefficients, and validation systems score the fitted equations; no system is reused across roles.

After column normalization, sequential thresholded least squares (STLSQ) applied to generation system $r$ at threshold $\lambda\in\Lambda$ produces support $\Sset_{r,\lambda}$ \citep{brunton2016discovering}. Deduplicating these supports gives the candidate set $\Cset$. Following stability selection and ensemble sparse discovery \citep{meinshausen2010stability,fasel2022ensemble}, we measure term recurrence by
\begin{equation}
  \begin{aligned}
  \pi_j
  &=\frac{1}{R_g|\Lambda|}
    \sum_{r=1}^{R_g}\sum_{\lambda\in\Lambda}
    \mathbf{1}\{j\in\Sset_{r,\lambda}\},\\
  \mathcal{J}_\tau
  &=\{j:\pi_j\ge\tau\},
  \qquad
  \Cset_\tau
  =\{\Sset\in\Cset:\Sset\subseteq\mathcal{J}_\tau\}.
  \end{aligned}
  \label{eq:term_stability}
\end{equation}
Thus $\pi_j$ is the selection frequency of term $j$, and $\Cset_\tau$ retains supports whose active terms all recur with frequency at least $\tau$. Using a threshold path exposes supports across multiple sparsity levels and reduces dependence on any single STLSQ threshold. The recurrence filter then excludes supports containing terms that do not recur across phases and thresholds.

Let $D_{\mathrm{fit}}$ contain the fit-system column norms and define
$\widetilde A_{\mathrm{fit}}=A_{\mathrm{fit}}D_{\mathrm{fit}}^{-1}$.
Each eligible support is then refit on the independent fit system by ridge regression:
\begin{equation}
  \begin{aligned}
  \hat{\widetilde\xi}_{\Sset}
  &=
  \arg\min_{\widetilde\xi:\mathrm{supp}(\widetilde\xi)\subseteq\Sset}
  \|\widetilde A_{\mathrm{fit}}\widetilde\xi-b_{\mathrm{fit}}\|_2^2
  +\alpha\|\widetilde\xi_{\Sset}\|_2^2,\\
  \hat\xi_{\Sset}&=D_{\mathrm{fit}}^{-1}\hat{\widetilde\xi}_{\Sset}.
  \end{aligned}
  \label{eq:ridge_refit}
\end{equation}
The second line restores the coefficients to the original library scale. We then compute the normalized residual on each validation system, with $\varepsilon=10^{-12}$ for numerical stability:
\begin{equation}
  \begin{aligned}
  r_s(\Sset)
  &=
  \frac{\|A_{\mathrm{val}}^{s}\hat\xi_{\Sset}-b_{\mathrm{val}}^{s}\|_2^2}
  {\|b_{\mathrm{val}}^{s}\|_2^2+\varepsilon},\\
  \mu(\Sset)
  &=\frac{1}{R_v}\sum_{s=1}^{R_v}r_s(\Sset).
  \end{aligned}
  \label{eq:validation_score}
\end{equation}
Let $\mathrm{se}(\Sset)$ denote the standard error across these validation risks, and let $\Sset_{\min}$ minimize their mean $\mu(\Sset)$. The one-standard-error rule \citep{hastie2009elements} admits
\begin{equation}
\begin{aligned}
  \Cset_{\mathrm{adm}}
  &=
  \{\Sset\in\Cset_\tau:\,
  \mu(\Sset)\le
  \mu(\Sset_{\min})+\mathrm{se}(\Sset_{\min})\},\\
  \widehat{\Sset}
  &=\arg\min_{\Sset\in\Cset_{\mathrm{adm}}}
  \bigl(|\Sset|,\mu(\Sset),-v(\Sset)\bigr).
\end{aligned}
  \label{eq:one_se_select}
\end{equation}
The tuple is ordered lexicographically, and $v(\Sset)$ is the number of times $\Sset$ is generated. SVWS therefore selects the smallest support whose independently refitted equation lies within one standard error of the minimum validation risk; lower risk and higher generation count break ties. The reported coefficients are $\hat\xi_{\widehat{\Sset}}$.

\noindent\textbf{Symbolic selection beyond fixed libraries.}
The proposal stage can generate symbolic expressions instead of fixed-library supports while retaining the frozen reconstruction, independent refitting, and validation on held-out weak systems. We consider
\begin{equation}
  u_t \approx \sum_{r=1}^{K} a_r\,\mathcal{D}_{\alpha_r}q_r(u),
  \qquad
  q_r\in\Gset,\quad
  \mathcal{D}_{\alpha_r}\in\mathcal{A},
  \label{eq:symbolic_operator_form}
\end{equation}
where $\Gset$ is the expression class generated by a symbolic grammar and the operators in $\mathcal A$ have known weak adjoints. A proposed pair $(q_r,\mathcal D_{\alpha_r})$ contributes the weak column $\langle q_r(\hat u),\mathcal D_{\alpha_r}^{*}\psi_\ell\rangle_\ell$. Integration by parts therefore transfers the outer differential operator to the test function and avoids pointwise differentiation of $q_r(\hat u)$.

Independent GP searches on weak systems built from shifted patch grids propose a finite set of expressions \citep{koza1992genetic,cranmer2023interpretable}. Because GP can produce many trees that share the same canonical structure but differ in their numerical parameters, each search groups them into expression families and retains the representative with the lowest proposal risk from each family. We merge these representatives and freeze the candidate pool before fitting their continuous parameters on an independent weak system. The fitted expressions are then compared on held-out weak systems using the mean validation risk and a fixed complexity penalty.

Our experiment uses $K=1$ and $\mathcal A=\{\partial_{xx}\}$, giving $u_t=\partial_{xx}q(u)$ after absorbing $a_1$ into $q$. The outer operator is therefore fixed, while the nonlinear diffusion function $q$ is discovered symbolically.

\section{Experiments}

\noindent\textbf{Benchmarks and observation protocols.}
We evaluate on public MDBench trajectories for three PDEs \citep{bideh2026mdbench}:
\begin{equation}
\begin{aligned}
  \text{KdV:}\quad
  u_t &= -6uu_x-u_{xxx},\\
  \text{KS:}\quad
  u_t &= -uu_x-u_{xx}-u_{xxxx},\\
  \text{2D AD:}\quad
  u_t &= 0.25u_x+0.5u_y+0.5u_{xx}+0.5u_{yy}.
\end{aligned}
\end{equation}
We use the released grids, initial conditions, and trajectories without resimulation. The main fixed-library comparison uses clean observations under two 20\% sampling protocols. In S20, a random 20\% subset of spatial locations is fixed and observed at every time point. In T20, the initial frame and randomly selected later frames form an irregular 20\% temporal subset observed on the full spatial grid. For each PDE and protocol, we evaluate five test seeds and provide every method with the same observations for a given seed. Candidate libraries are shared across methods: KdV and KS use the eight-term library $\{1,u,u^2,u^3,uu_x,u_{xx},u_{xxx},u_{xxxx}\}$, while 2D AD uses $\{1,u,u^2,u_x,u_y,u_{xx},u_{yy}\}$.

\begin{table*}[t]
  \centering
  {\small
  \setlength{\tabcolsep}{1.8pt}
  \begin{tabular*}{\textwidth}{@{\extracolsep{\fill}}lcccccc@{}}
    \toprule
    & \multicolumn{2}{c}{KdV} & \multicolumn{2}{c}{KS} & \multicolumn{2}{c}{2D AD} \\
    \cmidrule(lr){2-3}\cmidrule(lr){4-5}\cmidrule(lr){6-7}
    Method & S20 & T20 & S20 & T20 & S20 & T20 \\
    \midrule
    \textbf{Ours} & \textbf{5/5} (0.036) & \textbf{5/5} (0.008) & \textbf{5/5} (\underline{0.035}) & \textbf{5/5} (\underline{0.055}) & \textbf{5/5} (0.010) & \textbf{5/5} (\underline{0.009}) \\
    \addlinespace[1.5pt]
    PDE-FIND$^{\dagger}$ & 3/5 (0.044) & 4/5 (0.019) & 0/5 (0.136) & 3/5 (0.512) & 0/5 (0.894) & 0/5 (0.885) \\
    WSINDy$^{\dagger}$ & \textbf{5/5} (0.002) & 4/5 (0.011) & 1/5 (0.782) & 1/5 (0.818) & 0/5 (0.248) & \textbf{5/5} (0.099) \\
    \addlinespace[1.5pt]
    DeepMoD & \textbf{5/5} (\underline{0.001}) & \textbf{5/5} (\underline{0.001}) & 0/5 (1.010) & 0/5 (0.993) & \textbf{5/5} (\underline{0.008}) & \textbf{5/5} (0.012) \\
    PINN-SR$^{\ddagger}$ & \textbf{5/5} (0.015) & \textbf{5/5} (0.018) & 0/5 (0.998) & 0/5 (0.997) & 0/5 (0.326) & 0/5 (0.308) \\
    Weak-PDE-LEARN & 0/5 (0.326) & 0/5 (0.323) & 0/5 (1.238) & 0/5 (1.752) & 0/5 (0.675) & 0/5 (0.682) \\
    DL-PDE$^{\ddagger}$ & \textbf{5/5} (0.049) & \textbf{5/5} (0.094) & 3/5 (0.927) & 2/5 (0.905) & \textbf{5/5} (0.031) & \textbf{5/5} (0.018) \\
    \bottomrule
  \end{tabular*}
  \vspace{1pt}
  \parbox{\textwidth}{\small $^{\dagger}$Applied after fitting a shared spectral reconstruction to the sampled observations by ridge regression. $^{\ddagger}$Reimplemented from the published method.}
  }
  \caption{Fixed-library discovery from sparse observations. Cells report exact-support recovery (/5), with median relative coefficient error $E_\xi$ in parentheses. S20 observes 20\% fixed sensors; T20 observes 20\% time frames. Bold marks the highest recovery; underlining marks the lowest $E_\xi$ among ties.}
  \label{tab:fixed_library_main}
\end{table*}

\noindent\textbf{Baselines and implementation details.}
Ours combines the structured field adapter with SVWS. We compare with PDE-FIND, WSINDy, DeepMoD, PINN-SR, Weak-PDE-LEARN, and DL-PDE \citep{rudy2017data,messenger2021weak,both2021deepmod,chen2021physics,stephany2024weak,xu2021dlpde}. PDE-FIND and WSINDy operate on a shared spectral reconstruction fitted by ridge regression to the sampled observations. Hyperparameters not specified by the original sources are selected on disjoint development seeds and fixed across all test settings. Implementation sources and reporting rules are documented in Supplementary Sec.~C.

In the fixed-library experiments, SVWS uses 12 generation systems, one fit system, and seven validation systems. All adapter and selector settings are fixed before testing; their complete configurations are provided in Supplementary Secs.~A.3 and B.2.

\noindent\textbf{Metrics.}
Exact recovery requires $\widehat{\Sset}=\Sset^\star$ and is reported as the number of successful runs out of five. We also report relative coefficient and field errors:
\[
  E_\xi=\frac{\|\hat\xi-\xi^\star\|_2}{\|\xi^\star\|_2+\varepsilon},
  \qquad
  E_u=\frac{\|\hat u-u^\star\|_{2,\mathcal{X}\times\mathcal{T}}}
  {\|u^\star\|_{2,\mathcal{X}\times\mathcal{T}}},
\]
where $\varepsilon=10^{-12}$. For $E_\xi$, each discovered equation is mapped to the shared physical library, with zero coefficients assigned to absent terms. Field error is evaluated on the full reference grid, and error summaries are medians over the five test runs.

\noindent\textbf{Fixed-library recovery.}
Table~\ref{tab:fixed_library_main} reports the primary comparison. Our method recovers the exact support in all five runs across all six regimes. On KS, the best baseline reaches 3/5 exact recoveries in each observation protocol, whereas our method recovers all five and also attains the lowest median coefficient error. Supplementary Sec.~G shows that our KS recovery remains exact across the evaluated sensor densities and that exact support is retained under 10\% observation noise on KS and 2D AD.

\begin{table*}[t]
  \centering
  {\small
  \setlength{\tabcolsep}{5.0pt}
  \begin{tabular*}{\textwidth}{@{\extracolsep{\fill}}lcccc@{}}
    \toprule
    \multicolumn{5}{@{}l@{}}{\textbf{(a) Adapter ablations on KS}} \\
    & \multicolumn{2}{c}{KS-S20} & \multicolumn{2}{c}{KS-T20} \\
    \cmidrule(lr){2-3}\cmidrule(lr){4-5}
    Adapter & Exact & $E_u$ & Exact & $E_u$ \\
    \midrule
    \textbf{Structured adapter} & \textbf{5/5} & \textbf{0.082} & \textbf{5/5} & \textbf{0.082} \\
    Structured adapter without Fourier features & 3/5 & 0.659 & 5/5 & 0.650 \\
    Coordinate MLP with Fourier features & 3/5 & 0.440 & 1/5 & 0.437 \\
    Coordinate MLP without Fourier features & 0/5 & 0.905 & 0/5 & 0.867 \\
    \addlinespace[2pt]
    \midrule
    \multicolumn{5}{@{}l@{}}{\textbf{(b) Robustness to the STLSQ threshold}} \\
    Selection & KS-S20 & KS-T20 & 2D AD-S20 & 2D AD-T20 \\
    \midrule
    \textbf{SVWS} & \textbf{5/5} & \textbf{5/5} & \textbf{5/5} & \textbf{5/5} \\
    STLSQ ($0.25\lambda_0$) & 1/5 & 3/5 & 5/5 & 5/5 \\
    STLSQ ($\lambda_0$) & 5/5 & 4/5 & 5/5 & 5/5 \\
    STLSQ ($2\lambda_0$) & 5/5 & 5/5 & 0/5 & 0/5 \\
    \bottomrule
  \end{tabular*}
  }
  \caption{Adapter ablations and STLSQ threshold robustness. Panel (a) compares adapters with matched parameter counts on KS; panel (b) compares SVWS with single-system STLSQ along the prespecified threshold path on the same frozen fields. $E_u$ is median relative field error.}
  \label{tab:structure_ablation_main}
\end{table*}

\noindent\textbf{Adapter ablations and threshold robustness.}
Panel (a) of Table~\ref{tab:structure_ablation_main} focuses on KS, where adapter choice produces clear differences in field error and support recovery; Supplementary Sec.~D.1 reports all six regimes. With matched parameter counts, the structured adapter is the only variant that recovers all five supports under both observation protocols and gives the lowest field error by a clear margin. Fourier features reduce reconstruction error for both the factorized adapter and the coordinate MLP, while the factorized representation remains more reliable for equation recovery.

Panel (b) asks whether one STLSQ threshold transfers across equations and observation protocols, with $\lambda_0$ denoting the benchmark-specific base threshold. Lower thresholds admit additional terms on KS, whereas the largest threshold removes governing terms in both 2D AD regimes. SVWS instead generates candidates over the full prespecified path and multiple weak systems, maintaining exact recovery across all four settings. Additional selector diagnostics are reported in Supplementary Sec.~D.2.

\begin{figure*}[!t]
\centering
\includegraphics[width=0.94\textwidth]{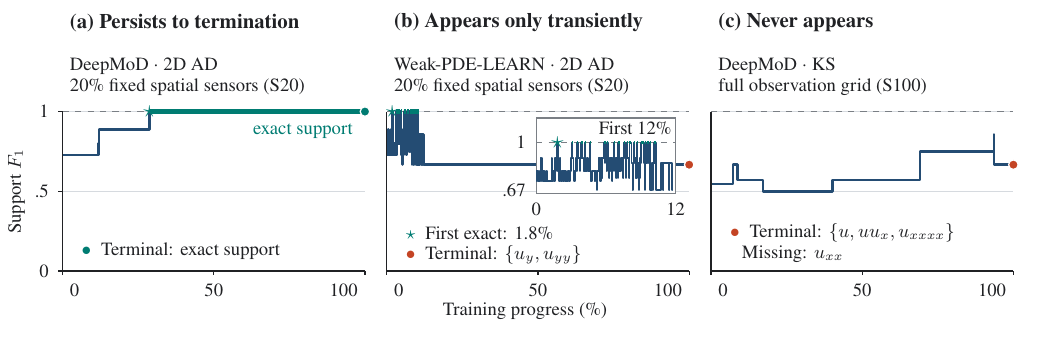}
\caption{Representative support trajectories under joint neural optimization. DeepMoD retains the exact 2D AD support on S20, Weak-PDE-LEARN visits it only transiently, and DeepMoD never visits the exact KS support on S100. Stars mark first exact visits and circles terminal states. Each reported behavior occurs in all five runs for its corresponding setting; complete paths are reported in Supplementary Sec.~E.}
\label{fig:joint_mechanism}
\end{figure*}

\noindent\textbf{Support dynamics under joint optimization.}
Figure~\ref{fig:joint_mechanism} traces support recovery along three coupled neural optimization paths. On 2D AD-S20, DeepMoD reaches the exact support and retains it to termination, whereas Weak-PDE-LEARN visits it only transiently during training. On KS-S100, DeepMoD never reaches the exact support at a recorded checkpoint despite observing the full spatial grid. These behaviors have different implications for checkpoint reporting: a transient visit makes the reported equation checkpoint-dependent, whereas an unvisited support cannot be recovered by changing the checkpoint. Freezing the field decouples equation selection from this path: candidates are refit on the same reconstruction and compared on held-out weak systems rather than selected from a training checkpoint.

\noindent\textbf{Symbolic discovery beyond a fixed library.}
We next examine whether the same freeze-then-select design supports symbolic discovery beyond a fixed library. The structured adapter is trained using observations from 20\% of the spatial sensors, after which genetic programming (GP) proposes candidate expressions and SVWS selects among them using independent weak systems. We consider the nonlinear diffusion equation

\[
  u_t=\partial_{xx}q(u),\qquad q^\star(u)=0.1u^{1.73},
\]
under 0\%, 5\%, or 10\% observation noise. Here $q$ is the unknown nonlinear diffusion function. Since $\partial_{xx}$ eliminates additive constants, $q$ and $q+C$ define the same PDE. The target exponent 1.73 is absent from the initialization grid and must be reached through mutation and continuous refitting.

We run two GP proposal searches on independently shifted weak systems. Each search groups candidates by canonical structure, and the retained representatives are merged into a frozen pool. A separate weak system fits their continuous constants and exponents, while three held-out systems evaluate the fitted expressions. Final selection combines the mean validation risk with a fixed complexity penalty. A run counts as power-law recovery when the selected expression canonicalizes to $\hat q(u)=\hat\kappa u^{\hat m}+C$. Table~\ref{tab:openform_capability} reports recovery frequency together with the corresponding parameter estimates. Supplementary Sec.~F documents the complete protocol.

PySR \citep{cranmer2023interpretable} does not reconstruct a field from sparse PDE observations, so we apply it to a weak-form estimate of $q(u)$ from the full noisy grid.

\begin{table}[t]
  \centering
  {\footnotesize
  \setlength{\tabcolsep}{1.5pt}
  \renewcommand{\arraystretch}{1.08}
  \begin{tabular*}{\columnwidth}{@{\extracolsep{\fill}}cccc@{}}
    \toprule
    \shortstack{Noise\\(\%)} & \shortstack{Power-law\\recovery (/5)} & \multicolumn{2}{c}{Parameters on recovered forms} \\
    \cmidrule(lr){3-4}
    & & $\hat\kappa$ & $\hat m$ \\
    \midrule
    \multicolumn{4}{@{}l}{\textbf{SVWS (ours)} \quad \textit{20\% fixed sensors}} \\
    0 & \textbf{5/5} & $0.0978\pm0.0019$ & $1.758\pm0.027$ \\
    5 & \textbf{5/5} & $0.0976\pm0.0026$ & $1.682\pm0.116$ \\
    10 & \textbf{5/5} & $0.0982\pm0.0041$ & $1.552\pm0.092$ \\
    \addlinespace[2pt]
    \multicolumn{4}{@{}l}{\textbf{PySR} \quad \textit{full-grid $q(u)$ estimate}} \\
    0 & 1/5 & $0.1000$ & $1.730$ \\
    5 & 3/5 & $0.0995\pm0.0005$ & $1.761\pm0.043$ \\
    10 & 2/5 & $0.0988\pm0.0014$ & $1.740\pm0.039$ \\
    \bottomrule
  \end{tabular*}
  }
  \caption{Symbolic recovery of $q^\star(u)=0.1u^{1.73}$ (five runs per noise level). SVWS uses 20\% fixed sensors; PySR is applied to a full-grid weak-form estimate of $q(u)$. Parameters are mean $\pm$ sample SD over successful recoveries.}
  \label{tab:openform_capability}
\end{table}

SVWS consistently recovers the target power-law structure from sparse sensor observations across all tested noise levels. PySR yields accurate parameter estimates for its recovered power laws, but selects the target structure less consistently from the full-grid $q(u)$ estimate. The comparison distinguishes accurate parameter fitting from reliable structural recovery and shows how the frozen field interface connects sparse observations to symbolic search.

\FloatBarrier

\section{Discussion}

The fixed-library experiments show complementary contributions from field reconstruction and equation selection. On KS, the structured adapter improves both field accuracy and support recovery. Across KS and two-dimensional advection--diffusion, single-system STLSQ is sensitive to the sparsity threshold, while SVWS remains consistent on the same frozen fields (Table~\ref{tab:structure_ablation_main}). In the external comparison, KdV and two-dimensional advection--diffusion are recovered by several baselines, whereas KS produces the clearest separation in support recovery and coefficient accuracy (Table~\ref{tab:fixed_library_main}). Together, these results show the value of combining structured reconstruction with selection across multiple weak systems and sparsity thresholds.

Figure~\ref{fig:joint_mechanism} provides further motivation for freezing. In coupled neural discovery, the reported equation depends on whether optimization reaches the correct support and whether that support survives to the selected checkpoint. A transient visit makes the result checkpoint-dependent, while a support absent from the recorded path cannot be recovered through checkpoint selection. The KS-S100 case further shows that complete spatial coverage does not ensure that the recorded path contains the correct support. The freeze-then-select design avoids this checkpoint dependence by refitting candidate equations on the same reconstruction and comparing them on held-out weak systems.

The symbolic experiment extends the same design from fixed-library supports to expressions proposed by GP. After the structured adapter converts sparse sensor observations into a frozen field, continuous parameters are fitted on one weak system and the candidate expressions are compared on held-out systems. SVWS consistently recovers the target power-law form, whereas PySR selects it less reliably from the full-grid $q(u)$ estimate (Table~\ref{tab:openform_capability}). By fixing the outer operator $\partial_{xx}$, the experiment isolates discovery of the nonlinear function and directly tests whether validation on the frozen field can extend beyond fixed-library support selection. The result supports this extension and motivates richer symbolic grammars that also search over multiple functional terms and differential operators.

\section{Conclusion}

We presented a freeze-then-select method for PDE discovery from sparse observations. A structured field adapter reconstructs a continuous field without a PDE residual, and SVWS uses independent weak systems for support generation, coefficient refitting, and validation. The method recovers exact support across all six sparse MDBench regimes, with its clearest gains on KS; its symbolic extension recovers the power-law form of an unknown nonlinear diffusion function from sparse, noisy observations. Optimization-path analysis further shows that coupled neural discovery may retain, lose, or never reach the exact support, motivating equation selection after reconstruction. By freezing the field before selection, the framework connects sparse measurements to both fixed-library and symbolic PDE discovery through a common validation principle.

\section*{Acknowledgments}

This work was supported by the Science Fund for Excellent Research Groups
of the National Natural Science Foundation of China (Grant No.~62588101)
and by the Science and Education Integration Project of the Shanghai
Institute of Technical Physics, CAS (Project
No.~SITPKJRH-2025-03).

\noindent\textbf{Code availability.}
Code and configurations will be made publicly available upon acceptance.

\bibliography{references}

\clearpage
\begingroup
\setcounter{secnumdepth}{2}
\setcounter{section}{0}
\setcounter{subsection}{0}
\setcounter{equation}{0}
\setcounter{figure}{0}
\setcounter{table}{0}
\renewcommand{\thesection}{\Alph{section}}
\renewcommand{\thesubsection}{\thesection.\arabic{subsection}}
\renewcommand{\theequation}{S\arabic{equation}}
\renewcommand{\thefigure}{S\arabic{figure}}
\renewcommand{\thetable}{S\arabic{table}}
\renewcommand{\thealgorithm}{S\arabic{algorithm}}
\makeatletter
\setlength{\@fptop}{0pt}
\setlength{\@fpsep}{8pt plus 2pt minus 2pt}
\setlength{\@fpbot}{0pt plus 1fil}
\setlength{\@dblfptop}{0pt}
\setlength{\@dblfpsep}{8pt plus 2pt minus 2pt}
\setlength{\@dblfpbot}{0pt plus 1fil}
\makeatother
\setcounter{topnumber}{4}
\setcounter{dbltopnumber}{4}
\renewcommand{\topfraction}{0.95}
\renewcommand{\dbltopfraction}{0.95}
\renewcommand{\textfraction}{0.05}
\renewcommand{\floatpagefraction}{0.75}
\renewcommand{\dblfloatpagefraction}{0.75}
\setlength{\floatsep}{8pt plus 2pt minus 2pt}
\setlength{\dblfloatsep}{8pt plus 2pt minus 2pt}
\setlength{\textfloatsep}{10pt plus 2pt minus 2pt}
\setlength{\dbltextfloatsep}{10pt plus 2pt minus 2pt}

\twocolumn[{\begin{center}
{\Large\bfseries Supplementary Material}\\[3pt]
{\large Freeze, Then Select: Structured Field Adapters and\\
Stability-Validated Weak Selection for PDE Discovery from Sparse Observations}
\end{center}
\vspace{6pt}}]

This supplement provides implementation details and additional results for
the experiments in the main paper. Sections A--C specify the experimental
protocols, method configuration, and baseline implementations; Sections D--G
present additional ablations, analyses of recorded optimization paths,
symbolic discovery details, and analyses of observation density and noise.
Section H summarizes the planned code release and computational resources.

\section{Experimental Details}

The fixed-library experiments use all five test seeds listed in
Sec.~A.1. Dense reference fields and ground-truth equations are used only for
evaluation.

\subsection{Benchmarks and Sparse Observation Protocols}
\label{sec:supp_benchmarks}

We use the public MDBench trajectories for KdV, Kuramoto--Sivashinsky (KS),
and two-dimensional advection--diffusion (2D AD). For each benchmark, we use
the released coordinates, initial condition, and the first released trajectory
(state index 0) without resimulation. Table~\ref{tab:benchmark_protocols}
summarizes the complete grids and the resulting observation counts under the
two sparse observation protocols.

\begin{table*}[t]
  \centering
  \small
  \setlength{\tabcolsep}{5pt}
  \begin{tabular*}{\textwidth}{@{\extracolsep{\fill}}lccccc@{}}
    \toprule
    PDE & Spatial grid & Time frames & Space--time domain & S20 sensors & T20 frames \\
    \midrule
    KdV & 512 & 201 & $[-30,30)\times[0,20]$ & 102 & 40 \\
    KS & 1024 & 251 & $[0,32\pi]\times[0,100]$ & 204 & 50 \\
    2D AD & $51\times51$ & 61 & $[-5,5]^2\times[0,6]$ & 520 & 12 \\
    \bottomrule
  \end{tabular*}
  \caption{Benchmark grids and observation counts. S20 reports the number of
  fixed spatial sensors observed at every time frame; T20 reports the number
  of retained frames observed on the complete spatial grid.}
  \label{tab:benchmark_protocols}
\end{table*}

The primary fixed-library comparison uses clean observations under two
complementary 20\% sampling protocols. In S20, each seed samples spatial
locations uniformly without replacement and observes the same sensor set at
every time frame. In T20, the initial frame is retained and the remaining
frames are sampled uniformly without replacement to form an irregular
temporal subset observed on the complete spatial grid. We evaluate the five
test seeds $\{1301,1709,2203,2917,3571\}$ for every benchmark and protocol.

For each benchmark, protocol, and seed, a shared observation record stores the
sampled coordinates and values, the spatial masks, and the retained time
indices. Every method receives the same record. Methods that fit continuous
surrogates may evaluate those surrogates at additional coordinates for
automatic differentiation or weak integration, but only the sampled values
enter model fitting. Dense reference values are used exclusively to compute
evaluation metrics.

The shared physical candidate libraries are
\begin{align}
\Theta_{\mathrm{KdV/KS}}
  &= [1,u,u^2,u^3,u u_x,u_{xx},u_{xxx},u_{xxxx}],\\
\Theta_{\mathrm{2D\,AD}}
  &= [1,u,u^2,u_x,u_y,u_{xx},u_{yy}].
\end{align}
These physical libraries are fixed before evaluation and shared by all
methods. Algebraically equivalent conservative terms are mapped to the
displayed convention before support and coefficient scoring; for example,
$\partial_x(u^2)=2u u_x$.

\subsection{Metrics and Aggregation}

We evaluate structural recovery separately from coefficient and field
accuracy. Exact recovery requires equality between the selected support
$\widehat{\Sset}$ and the true support $\Sset^\star$. We define support
$F_1$ as the harmonic mean of precision and recall over candidate terms. The
relative coefficient and field errors are
\begin{equation}
\begin{aligned}
 E_\xi
 &=\frac{\norm{\hat\xi-\xi^\star}_2}
 {\norm{\xi^\star}_2+10^{-12}},\\
 E_u
 &=\frac{\norm{\hat u-u^\star}_2}
 {\norm{u^\star}_2}.
\end{aligned}
\end{equation}
Before computing $E_\xi$, each selected equation is expressed in the shared
physical basis defined in Sec.~A.1, with zero coefficients assigned to absent
terms. Field error is evaluated on the complete reference grid.

The fixed library results report exact recoveries out of five runs and medians
over the same five runs. Curves in the observation density study report
medians and interquartile ranges. In the symbolic experiment, continuous
parameters are summarized as mean $\pm$ sample standard deviation over runs
that recover the target power-law structure.

\subsection{Structured Field Adapter and Training}
\label{sec:supp_adapter_training}

The structured field adapter reconstructs the continuous field as
\begin{equation}
  \widehat u_{\eta,C}(x,t)
  =b_\eta(x)+\Phi_\eta(x)^\top C\bar\beta(t)
\end{equation}
using only the sampled observations. A spatial MLP jointly produces the
background $b_\eta(x)$ and $R$ spatial features $\Phi_\eta(x)$. Its input is
the random Fourier encoding
$\gamma(\tilde x)=[\sin(B\tilde x),\cos(B\tilde x)]$, where the 64 rows of $B$
are sampled once per run from a standard Gaussian distribution and then held
fixed. The hidden layers use Swish activations. Here $\tilde x=x$ for KdV and
KS, while each coordinate of 2D AD is divided by 5 before Fourier encoding.
The temporal vector $\bar\beta(t)$ is a clamped uniform cubic B-spline basis
centered by subtracting its empirical mean over the observed time locations,
with $K$ internal knots and $M=K+4$ basis functions. The matrix
$C\in\mathbb{R}^{R\times M}$ contains the trainable spline coefficients. S20
and T20 use the same adapter configuration; they differ only in the
coordinates and values supplied to the reconstruction loss.
Table~\ref{tab:adapter_configs} lists the benchmark-specific architecture
sizes and maximum training epochs.

\begin{table*}[t]
  \centering
  \small
  \setlength{\tabcolsep}{4.2pt}
  \begin{tabular*}{\textwidth}{@{\extracolsep{\fill}}lcccccc@{}}
    \toprule
    PDE & Rank $R$ & Hidden width $\times$ layers & Internal knots $K$ &
    Basis size $M$ & Trainable parameters & Maximum epochs \\
    \midrule
    KdV & 12 & $64\!\times\!3$ & 32 & 36 & 17,853 & 5,000 \\
    KS & 24 & $64\!\times\!3$ & 64 & 68 & 19,833 & 5,500 \\
    2D AD & 16 & $72\!\times\!3$ & 6 & 10 & 21,201 & 5,500 \\
    \bottomrule
  \end{tabular*}
  \caption{Structured adapter settings. All models use Swish activations and
  the random Fourier encoding described above. Trainable parameter counts
  include the spatial MLP and $C$.}
  \label{tab:adapter_configs}
\end{table*}

The adapter is trained from observations without a PDE residual:
\begin{equation}
  \min_{\eta,C}\;
  \mathcal{L}_{\mathrm{obs}}
  +\lambda_\Phi\mathcal{R}_\Phi
  +\lambda_t\mathcal{R}_t,
\end{equation}
where
\begin{align}
  \mathcal{L}_{\mathrm{obs}}
  &=\frac{1}{N_{\mathrm{obs}}}
    \sum_i\left|
    \widehat u_{\eta,C}(x_i,t_i)-y_i
    \right|^2,\\
  \mathcal{R}_\Phi
  &= \left\|
     \frac{1}{N_g}\Phi_\eta(X_g)^\top\Phi_\eta(X_g)-I
     \right\|_F^2,\\
  \mathcal{R}_t
  &= \sum_{r=1}^{R}\int_0^T |c_r''(t)|^2\,dt,
\end{align}
where $X_g$ contains $N_g$ points sampled uniformly from the spatial domain
and $c_r(t)=C_{r:}\bar\beta(t)$ is the coefficient of spatial feature $r$.
The Gram penalty $\mathcal{R}_\Phi$ uses 512 uniformly sampled spatial points
per update, and the temporal curvature integral is approximated on a uniform
200-point time grid.

Training uses AdamW with an initial learning rate of $10^{-3}$, a cosine
decay schedule over the maximum epoch budget, weight decay $10^{-2}$, and
global gradient-norm clipping at 1.0. During the first 400 epochs, $C$ is
fixed at zero to fit the background component. Afterward, $\eta$ and $C$ are
optimized jointly. For KdV and KS, both regularization weights are zero for
the first 800 epochs and increase linearly over the next 1,600 epochs; for 2D
AD, the corresponding periods are 600 and 1,200 epochs. Their target weights
are $\lambda_\Phi=3\times10^{-3}$ and $\lambda_t=5\times10^{-4}$. The weighted
feature and temporal penalties are capped at 0.15 and 0.25 times the current
observation loss, respectively. Each update samples at most 32 observed time
frames and uses all spatial observations available at those frames. The MSE
over all observations selects the lowest-loss reconstruction checkpoint among
those tracked after the 400-epoch background phase. Early stopping begins at
the same point and uses a patience of 800 epochs and a minimum decrease of
$10^{-6}$. The selected checkpoint is frozen before any weak system is
constructed; equation selection does not feed back into adapter training.

\section{Stability-Validated Weak Selection}

\subsection{Weak System Construction and Continuity}
\label{sec:supp_weak_systems}

Consider a candidate equation written as
\begin{equation}
  u_t=\sum_j\xi_j\mathcal{D}_{\alpha_j}q_j(u),
\end{equation}
where $\alpha_j\in\mathbb{N}_0^d$,
$\mathcal{D}_{\alpha_j}=\partial_x^{\alpha_j}$, and $q_j$ is a scalar
function of the state. The case $\alpha_j=0$ covers algebraic terms, and the
formal adjoint is
$\mathcal{D}_{\alpha_j}^{*}=(-1)^{|\alpha_j|}
\partial_x^{\alpha_j}$. For a compactly supported test function $\psi_\ell$,
integration by parts transfers the derivatives to $\psi_\ell$.
\begin{samepage}
The corresponding weak entries are
\begin{equation}
\begin{aligned}
  b_\ell(\widehat u)
  &=-\langle\widehat u,\partial_t\psi_\ell\rangle_\ell,\\
  A_{\ell j}(\widehat u)
  &=\langle q_j(\widehat u),
  \mathcal{D}_{\alpha_j}^{*}\psi_\ell\rangle_\ell,
\end{aligned}
\end{equation}
\end{samepage}
where $\langle\cdot,\cdot\rangle_\ell$ denotes integration over patch
$P_\ell$. These entries form the regression system
$b(\widehat u)\approx A(\widehat u)\xi$. For a patch centered at $c_\ell$,
with spatial and temporal
half-widths $h_{\ell,a}$ and $h_{\ell,t}$, we use
\begin{align}
  \rho_p(s)
  &= (1-s^2)^p\mathbf{1}\{|s|<1\},\\
  \psi_\ell(x,t)
  &= \rho_p\!\left(\frac{t-c_{\ell,t}}{h_{\ell,t}}\right)
     \prod_{a=1}^{d}
     \rho_p\!\left(\frac{x_a-c_{\ell,a}}{h_{\ell,a}}\right).
\end{align}
Because $\rho_p$ and its derivatives through order $p-1$ vanish at
$s=\pm1$, choosing $p$ above the highest derivative order in each candidate
library eliminates the boundary terms introduced by integration by parts.
Spatial derivatives therefore act on the known test function rather than
pointwise on the reconstructed field. This construction uses values from the
frozen adapter while avoiding direct evaluation of its higher-order spatial
derivatives.

Patch centers form a regular grid inside the admissible domain. Each weak
system applies an independently sampled phase shift to this grid. The field,
observations, and candidate library remain unchanged; only the integration
patches differ. This produces multiple views of the same frozen reconstruction
and provides the variation used by SVWS to assess support stability. All
integrals are evaluated using midpoint tensor grids.
Table~\ref{tab:weak_system_configs} lists the corresponding settings for each
benchmark.

\begin{table*}[t]
  \centering
  \small
  \setlength{\tabcolsep}{5pt}
  \begin{tabular*}{\textwidth}{@{\extracolsep{\fill}}lccccc@{}}
    \toprule
    PDE & Patches per system & Nodes per patch & Spatial half-width &
    Temporal half-width & Kernel power $p$ \\
    \midrule
    KdV & 300 & 3,000 & $4.8$ & $1.0$ & 8 \\
    KS & 300 & 3,100 & $8.0$ & $1.0$ & 8 \\
    2D AD & 320 & 2,280 & $(1.6,1.6)$ & $0.48$ & 4 \\
    \bottomrule
  \end{tabular*}
  \caption{Weak system settings for each benchmark. Nodes per patch give the
  effective size of the midpoint tensor grid; all half-widths are in physical
  coordinates.}
  \label{tab:weak_system_configs}
\end{table*}

For the exact weak entries, suppose that $u,\widehat u\in L_2(P_\ell)$, the
required derivatives of $\psi_\ell$ lie in $L_2(P_\ell)$, and $q_j$ is
$L_j$-Lipschitz on an interval containing the essential ranges of $u$ and
$\widehat u$ over $P_\ell$. Then
\begin{align}
  |b_\ell(\widehat u)-b_\ell(u)|
  &\leq
  \|\widehat u-u\|_{L_2(P_\ell)}
  \|\partial_t\psi_\ell\|_{L_2(P_\ell)},\\
  |A_{\ell j}(\widehat u)-A_{\ell j}(u)|
  &\leq
  L_j\|\widehat u-u\|_{L_2(P_\ell)} \notag\\[-2pt]
  &\quad\times
  \|\mathcal D_{\alpha_j}^{*}\psi_\ell\|_{L_2(P_\ell)}.
\end{align}
These bounds follow from the Cauchy--Schwarz inequality and establish
entrywise continuity of the exact weak system in the local reconstruction
error under the stated range condition. The midpoint discretization satisfies
analogous bounds in the corresponding quadrature-weighted norm. Comparing
independently shifted systems then allows SVWS to favor supports that remain
stable across weak discretizations. Support recovery also depends on library
conditioning, trajectory excitation, and coefficient separation.

\subsection{Selection Algorithm and Configuration}
\label{sec:supp_svws_algorithm}

\begin{figure*}[t]
\refstepcounter{algorithm}\label{alg:svws}
\hrule
\noindent Algorithm~\thealgorithm: Stability-Validated Weak Selection on a frozen field
\par\hrule
\footnotesize
\begin{tabularx}{\linewidth}{@{}r@{\hspace{5pt}}Y@{}}
\multicolumn{2}{@{}p{\linewidth}@{}}{\textbf{Input:} frozen field $\hat u$; candidate library; base threshold $\lambda_0$; multipliers $\mathcal{M}$; system counts $R_g,R_v$; independent phase seeds; stability threshold $\tau$; ridge parameter $\alpha$; numerical constant $\varepsilon$.}\\[2pt]
1 & Build independently phased weak systems $\{(A_{\mathrm{gen}}^r,b_{\mathrm{gen}}^r)\}_{r=1}^{R_g}$, $(A_{\mathrm{fit}},b_{\mathrm{fit}})$, and $\{(A_{\mathrm{val}}^s,b_{\mathrm{val}}^s)\}_{s=1}^{R_v}$ from the same frozen field.\\
2 & For each $(r,m)\in\{1,\ldots,R_g\}\times\mathcal{M}$, run STLSQ at threshold $m\lambda_0$ on the column-normalized generation system and record support $\Sset_{r,m}$.\\
3 & Deduplicate the supports to form $\Cset$. For each $\Sset\in\Cset$, record its generation count $v(\Sset)=\sum_{r,m}\mathbf{1}\{\Sset_{r,m}=\Sset\}$. Compute term frequencies $\pi_j=(R_g|\mathcal{M}|)^{-1}\sum_{r,m}\mathbf{1}\{j\in\Sset_{r,m}\}$ and stable terms $\mathcal{J}_\tau=\{j:\pi_j\ge\tau\}$.\\
4 & Retain $\Cset_\tau=\{\Sset\in\Cset:\Sset\subseteq\mathcal{J}_\tau\}$. If this set is empty, set $\Cset_\tau=\Cset$.\\
5 & Let $D_{\mathrm{fit}}$ contain the fit-system column norms and set $\widetilde A_{\mathrm{fit}}=A_{\mathrm{fit}}D_{\mathrm{fit}}^{-1}$. For every $\Sset\in\Cset_\tau$, fit $\widetilde\xi_{\Sset}$ by ridge regression on $(\widetilde A_{\mathrm{fit}},b_{\mathrm{fit}})$ and restore the original library scale with $\hat\xi_{\Sset}=D_{\mathrm{fit}}^{-1}\widetilde\xi_{\Sset}$.\\
6 & On each validation system, compute $r_s(\Sset)=\norm{A_{\mathrm{val}}^s\hat\xi_{\Sset}-b_{\mathrm{val}}^s}_2^2/(\norm{b_{\mathrm{val}}^s}_2^2+\varepsilon)$. Record the mean $\mu(\Sset)$ and standard error $\mathrm{se}(\Sset)$.\\
7 & Let $\Sset_{\min}\in\arg\min_{\Sset\in\Cset_\tau}\mu(\Sset)$ and admit every $\Sset$ satisfying $\mu(\Sset)\le\mu(\Sset_{\min})+\mathrm{se}(\Sset_{\min})$.\\
8 & Select the admissible support lexicographically by $(|\Sset|,\mu(\Sset),-v(\Sset))$.\\[2pt]
\multicolumn{2}{@{}p{\linewidth}@{}}{\textbf{Output:} selected support $\widehat{\Sset}$ and coefficients $\hat\xi_{\widehat{\Sset}}$ fitted on the independent fit system and mapped to the original library scale.}
\end{tabularx}
\hrule
\end{figure*}

Algorithm~\ref{alg:svws} summarizes SVWS. It assigns support generation,
coefficient fitting, and validation to disjoint weak systems built from the
same frozen field. Independently shifted patch grids provide different weak
views: generation systems produce candidate supports across sparsity levels,
the fit system estimates their physical coefficients, and validation systems
compare the fitted equations on held-out patches. All three stages operate
after the adapter is frozen.

We use $R_g=12$ generation systems and the threshold multipliers
\(
\{0.25,0.5,0.75,1,1.25,1.5,2\}
\)
around the base STLSQ threshold. The resulting $12\times7=84$ support
proposals are deduplicated, and term recurrence is computed over all
generation-system and threshold-multiplier pairs. Terms with recurrence at
least $\tau=0.5$ define the stable candidate pool.

Each retained support is refitted by ridge regression on one independent fit
system and evaluated on $R_v=7$ independently shifted validation systems. The
STLSQ solves and fixed-support refits both use ridge $10^{-4}$, with at most
20 STLSQ iterations; the normalized validation risk uses
$\varepsilon=10^{-12}$. Columns are normalized during STLSQ and coefficient
fitting, then mapped back to the physical library scale before validation. The
base thresholds for KdV, KS, and 2D AD are $0.2$, $0.4$, and $0.4$,
respectively. These settings were selected using development seeds
$\{42,505,606,808,909\}$, then shared across S20 and T20 and fixed for all test
seeds.

\section{External Baseline Implementations}

All baselines receive the sampled observations and physical candidate
libraries used in the main comparison. Each retains the optimization and
support selection procedure specified by its source implementation or
published algorithm. PDE-FIND and WSINDy operate on a shared spectral
reconstruction of the samples, whereas the neural methods use the observations
in their data objectives. Table~\ref{tab:baseline_implementation} identifies
the implementation, observation interface, and coefficient readout for each
method. Settings not specified by the original source are chosen on
development seeds and fixed before evaluation.

\begin{table*}[t]
  \centering
  \small
  \setlength{\tabcolsep}{3.2pt}
  \begin{tabularx}{\textwidth}{@{}P{0.14\textwidth}P{0.23\textwidth}P{0.34\textwidth}Y@{}}
    \toprule
    Method & Implementation & Observation interface & Reporting rule \\
    \midrule
    PDE-FIND & Reimplementation following the published procedure & Spectral reconstruction fitted to the sampled observations & STLSQ support; fixed-support ridge coefficients \\
    WSINDy & Reimplementation following the published weak-form procedure & Shared spectral reconstruction evaluated through weak systems & STLSQ support; coefficient refit on an independent weak system \\
    DeepMoD & Released DeePyMoD v2.2.0 implementation & Sampled observations enter the neural data objective & Terminal sparsity mask and unscaled constraint coefficients \\
    PINN-SR & Reimplementation of the published ADO procedure & Sampled observations enter the alternating data objective & Last STRidge projection from alternating optimization \\
    Weak-PDE-LEARN & Authors' released implementation with the Rational network & Sampled observations are supplied through its data interface & Final sparsification checkpoint \\
    DL-PDE & Reimplementation of the published two-stage procedure & A neural surrogate is fitted to the sampled observations & Terminal STRidge support; fixed-support ridge coefficients \\
    \bottomrule
  \end{tabularx}
  \caption{Implementations, observation interfaces, and reporting rules for
  the external baselines.}
  \label{tab:baseline_implementation}
\end{table*}

\subsection{Implementation Sources and Reference Experiments}

PDE-FIND and WSINDy are reimplemented following their published regression
procedures. DeepMoD uses DeePyMoD v2.2.0, and Weak-PDE-LEARN uses the authors'
released Rational network. For the released neural methods, benchmark wrappers
supply the MDBench observations and map native terms to the shared physical
library; optimization, sparsification, and terminal reporting follow the
released code. In particular, $D_x(u^2)$ in Weak-PDE-LEARN is mapped by
$D_x(u^2)=2u u_x$. Source revisions and dependencies are included in the code.

We reimplement PINN-SR and DL-PDE following the algorithms and training
procedures described in their papers. PINN-SR uses alternating direction
optimization. DL-PDE fits an observation-only neural field and applies STRidge
to quantities evaluated by automatic differentiation. Their complete
configurations are included in the code.

We also evaluate the released neural pipelines on their reference
configurations before introducing the MDBench interface and shared physical
library. Weak-PDE-LEARN recovers the exact KdV-Sine support in all five runs
using 4,000 measurements, 25\% noise, and the paper-specified schedule for the
Rational network. DeePyMoD recovers the exact support for its notebook KdV
example in two of five runs under the public v2.2.0 configuration. These
reference runs use each source's own data, library, and reporting convention;
they are distinct from the MDBench S20 and T20 experiments.

\subsection{Configurations and Parameter Selection}
\label{sec:supp_hyperparameters}

Each external method uses one PDE-specific configuration, shared across S20,
T20, and all five test seeds. PDE-FIND and WSINDy use ridge penalties of
$10^{-3}$ for spectral reconstruction and $10^{-4}$ for coefficient refitting.
DeepMoD uses a sparsification threshold of $0.1$, four hidden layers of width
30 for KdV and KS and width 50 for 2D AD, and at most 100,000 iterations.
Weak-PDE-LEARN uses five hidden layers of width 40 with Rational activations,
6,000 burn-in epochs, and 5,000 sparsification epochs. PINN-SR performs six
alternating optimization cycles. DL-PDE applies STRidge to the fixed terminal
surrogate checkpoint. Complete schedules are provided in the accompanying
configurations.

Table~\ref{tab:development_settings} reports the candidate values and selected
settings. The PDE-FIND, WSINDy, and DL-PDE thresholds are selected on seeds
$\{42,505,606,808,909\}$ using aggregate performance across S20 and T20.
We maximize exact recoveries, breaking ties by higher median support $F_1$,
lower median fixed-support coefficient error, and the smaller threshold. The
two KS adapter capacities are compared on independent development runs. Relative to
$(16,32)$, the selected $(24,64)$ configuration preserves exact recovery and
field accuracy under both protocols while reducing coefficient error. All
selection runs are disjoint from the reported test seeds.

\begin{table*}[t]
  \centering
  \small
  \setlength{\tabcolsep}{3pt}
  \begin{tabularx}{\textwidth}{@{}P{0.20\textwidth}P{0.56\textwidth}Y@{}}
    \toprule
    Hyperparameter & Candidate values & Selected value(s) \\
    \midrule
    KS adapter $(R,K)$
      & $(R,K)\in\{(16,32),(24,64)\}$
      & $(24,64)$ \\
    PDE-FIND threshold
      & $\{0.05,0.1,0.2,0.4,0.8,1,2,5\}$
      & $(0.2,5,0.05)$ \\
    WSINDy threshold
      & $\{0.05,0.1,0.2,0.4,0.8,1,2,5\}$
      & $(0.2,2,0.4)$ \\
    DL-PDE STRidge tolerance
      & $\{0.01,0.02,0.05,0.1,0.2,0.5,1,2,5,10,20,50,100,200,500,1000\}$
      & $(1,2,2)$ \\
    \bottomrule
  \end{tabularx}
  \caption{Hyperparameter candidate sets and selected values. For the three
  baseline rows, selected values are ordered as KdV, KS, and 2D AD; each
  PDE-specific choice is shared across S20 and T20.}
  \label{tab:development_settings}
\end{table*}

\section{Additional Ablations}

\subsection{Adapter Representation Across Benchmarks}

Table~\ref{tab:adapter_crossbenchmark_supp} extends the KS ablation in the
main paper to all six fixed-library settings. Within each PDE, the four
adapters have closely matched parameter counts and share the observations,
optimization budget, weak-form systems, and SVWS configuration. The
comparison isolates the effects of space--time factorization and fixed
spatial Fourier features.

\begin{table*}[t]
  \centering
  {\small
  \setlength{\tabcolsep}{3.8pt}
  \begin{tabular*}{\textwidth}{@{\extracolsep{\fill}}lcccccc@{}}
    \toprule
    & \multicolumn{2}{c}{KdV} & \multicolumn{2}{c}{KS} &
      \multicolumn{2}{c}{2D AD} \\
    \cmidrule(lr){2-3}\cmidrule(lr){4-5}\cmidrule(lr){6-7}
    Adapter & S20 & T20 & S20 & T20 & S20 & T20 \\
    \midrule
    \multicolumn{7}{@{}l@{}}{\textbf{(a) Exact support recovery}} \\
    \textbf{Structured adapter (ours)}
      & \textbf{5/5} & \textbf{5/5} & \textbf{5/5} & \textbf{5/5}
      & \textbf{5/5} & \textbf{5/5} \\
    Structured adapter without Fourier features
      & \textbf{5/5} & \textbf{5/5} & 3/5 & \textbf{5/5}
      & \textbf{5/5} & \textbf{5/5} \\
    Coordinate MLP with Fourier features
      & \textbf{5/5} & \textbf{5/5} & 3/5 & 1/5
      & \textbf{5/5} & \textbf{5/5} \\
    Coordinate MLP without Fourier features
      & 3/5 & 4/5 & 0/5 & 0/5 & \textbf{5/5} & \textbf{5/5} \\
    \addlinespace[2pt]
    \midrule
    \multicolumn{7}{@{}l@{}}{\textbf{(b) Median field error $E_u$}} \\
    \textbf{Structured adapter (ours)}
      & 0.1171 & 0.0239 & \textbf{0.0817} & \textbf{0.0815}
      & 0.0028 & \textbf{0.0026} \\
    Structured adapter without Fourier features
      & 0.1336 & 0.0979 & 0.6588 & 0.6503 & 0.0055 & 0.0046 \\
    Coordinate MLP with Fourier features
      & \textbf{0.0450} & \textbf{0.0236} & 0.4397 & 0.4365
      & \textbf{0.0026} & 0.0034 \\
    Coordinate MLP without Fourier features
      & 0.1699 & 0.0716 & 0.9046 & 0.8674 & 0.0139 & 0.0124 \\
    \bottomrule
  \end{tabular*}
  }
  \caption{Adapter ablations across all six fixed-library settings. Within
  each PDE, the four variants have closely matched parameter counts and share
  the observations, optimization budget, weak-form systems, and SVWS
  configuration. Panel (a) reports exact-support recoveries over five seeds,
  and panel (b) reports median relative field error. Bold marks the best value
  in each column, including ties.}
  \label{tab:adapter_crossbenchmark_supp}
\end{table*}

KdV and 2D AD are recovered exactly by several adapters and therefore provide
limited separation between representations. KS is more discriminating. The
structured adapter recovers all five supports under both protocols, whereas
the coordinate MLP with Fourier features recovers three under S20 and one
under T20, and the coordinate MLP without Fourier features recovers none.
Removing Fourier features from the structured adapter also substantially
increases its KS field error. Fourier encoding alone therefore does not
explain the recovery gain; its benefit is strongest when combined with the
structured separation of space and time.

Table~\ref{tab:adapter_large_control_supp} controls for coordinate MLP
capacity. Increasing the hidden width from 72 to 128 improves field accuracy
and exact recovery, yet the 49,665-parameter MLP remains less reliable than
the 19,833-parameter structured adapter.

\begin{table*}[t]
  \centering
  {\small
  \setlength{\tabcolsep}{4.2pt}
  \begin{tabular*}{\textwidth}{@{\extracolsep{\fill}}lrcccc@{}}
    \toprule
    & & \multicolumn{2}{c}{KS-S20} & \multicolumn{2}{c}{KS-T20} \\
    \cmidrule(lr){3-4}\cmidrule(lr){5-6}
    Adapter & Parameters & Exact & $E_u$ & Exact & $E_u$ \\
    \midrule
    \textbf{Structured adapter (ours)}
      & 19,833 & \textbf{5/5} & \textbf{0.082}
      & \textbf{5/5} & \textbf{0.082} \\
    Coordinate MLP with Fourier features (width 72)
      & 19,873 & 3/5 & 0.440 & 1/5 & 0.437 \\
    Coordinate MLP with Fourier features (width 128)
      & 49,665 & 4/5 & 0.340 & 3/5 & 0.315 \\
    \bottomrule
  \end{tabular*}
  }
  \caption{Capacity comparison on KS. The two coordinate MLPs differ only in
  hidden width; the structured adapter is included as a reference at a
  comparable parameter count. All rows use the same observations, training
  budget, and SVWS configuration. $E_u$ is the median relative field error
  over five seeds.}
  \label{tab:adapter_large_control_supp}
\end{table*}

\FloatBarrier
\subsection{Selector Components and Sensitivity}

Table~\ref{tab:selector_analysis_supp} evaluates selector sensitivity on KS
and 2D AD over the same five test seeds. Panel (a) uses the clean frozen fields
and configurations from the main comparison and extends the three-point
comparison to the complete seven-point threshold path. Each STLSQ row uses
the first generation weak system, so only the threshold changes.

\begin{table*}[t]
  \centering
  {\small
  \setlength{\tabcolsep}{5pt}
  \begin{tabular*}{\textwidth}{@{\extracolsep{\fill}}lcccc@{}}
    \toprule
    \multicolumn{5}{@{}l@{}}{\textbf{(a) Sensitivity to a fixed STLSQ threshold on clean frozen fields}} \\
    Selection & KS-S20 & KS-T20 & 2D AD-S20 & 2D AD-T20 \\
    \midrule
    \textbf{SVWS} & 5/5 & 5/5 & 5/5 & 5/5 \\
    STLSQ ($\lambda/\lambda_0=0.25$) & 1/5 & 3/5 & 5/5 & 5/5 \\
    STLSQ ($\lambda/\lambda_0=0.5$) & 4/5 & 4/5 & 5/5 & 5/5 \\
    STLSQ ($\lambda/\lambda_0=0.75$) & 5/5 & 4/5 & 5/5 & 5/5 \\
    STLSQ ($\lambda/\lambda_0=1$) & 5/5 & 4/5 & 5/5 & 5/5 \\
    STLSQ ($\lambda/\lambda_0=1.25$) & 5/5 & 4/5 & 5/5 & 5/5 \\
    STLSQ ($\lambda/\lambda_0=1.5$) & 5/5 & 5/5 & 2/5 & 2/5 \\
    STLSQ ($\lambda/\lambda_0=2$) & 5/5 & 5/5 & 0/5 & 0/5 \\
    \addlinespace[3pt]
    \midrule
    \multicolumn{5}{@{}l@{}}{\textbf{(b) Selector variants with 10\% observation noise}} \\
    Selection & KS-S20 & KS-T20 & 2D AD-S20 & 2D AD-T20 \\
    \midrule
    \textbf{SVWS} & 5/5 & 5/5 & 5/5 & 5/5 \\
    Without stability filtering & 5/5 & 5/5 & 4/5 & 5/5 \\
    Minimum mean validation risk & 1/5 & 1/5 & 3/5 & 5/5 \\
    Term-wise majority vote & 5/5 & 5/5 & 5/5 & 5/5 \\
    \bottomrule
  \end{tabular*}
  }
  \caption{Threshold and selector sensitivity on frozen fields. Panel (a)
  applies STLSQ to the first generation weak system along the prespecified
  threshold path on the clean fields from the main comparison. Panel (b)
  compares selector variants on fields reconstructed from observations with
  10\% noise. Entries are exact-support recoveries over five seeds.}
  \label{tab:selector_analysis_supp}
\end{table*}

Low thresholds retain extra KS terms, whereas high thresholds remove
governing terms from 2D AD. No single threshold recovers every run across all
four settings. SVWS instead considers supports along the complete path and
across multiple weak systems, recovering all five runs in every setting.

Panel (b) compares selector rules on fields reconstructed from observations
with 10\% noise. Within each run, all variants share the frozen field,
candidate pool, coefficient refits, and validation systems. Selecting the
minimum mean validation risk over the full pool frequently retains extra
terms, especially on KS. Removing stability filtering loses one 2D AD-S20
recovery. Term-wise majority voting matches SVWS in these four settings,
showing that recurrence alone can separate the correct terms once the field
is reliably reconstructed. SVWS achieves the same exact recovery while
combining recurrence filtering with held-out weak validation.

\subsection{Adapter Objective}

The adapter objective augments the observation loss with penalties that
encourage feature diversity and temporal smoothness. To isolate their
contribution, we compare the full objective with
$\lambda_\Phi=\lambda_t=0$. The variants use the same architecture,
optimizer, training budget, observations, and SVWS configuration, and both
retain AdamW weight decay. Table~\ref{tab:adapter_objective_ablation} reports
paired runs for all six S20 and T20 settings over the five test seeds.

\begin{table*}[t]
  \centering
  {\footnotesize
  \setlength{\tabcolsep}{3.0pt}
  \begin{tabular*}{\textwidth}{@{\extracolsep{\fill}}llcccccc@{}}
    \toprule
    & & \multicolumn{2}{c}{KdV} & \multicolumn{2}{c}{KS}
      & \multicolumn{2}{c}{2D AD} \\
    \cmidrule(lr){3-4}\cmidrule(lr){5-6}\cmidrule(lr){7-8}
    Objective & Metric & S20 & T20 & S20 & T20 & S20 & T20 \\
    \midrule
    Full objective & Exact recovery & 5/5 & 5/5 & 5/5 & 5/5 & 5/5 & 5/5 \\
      & Median $E_u$ & 0.117 & 0.024 & 0.082 & 0.082 & 0.003 & 0.003 \\
      & Maximum $E_u$ & 0.143 & 0.037 & 0.127 & 0.108 & 0.003 & 0.004 \\
    \addlinespace
    No explicit penalties & Exact recovery & 5/5 & 5/5 & 5/5 & 4/5 & 5/5 & 5/5 \\
      & Median $E_u$ & 0.116 & 0.030 & 0.073 & 0.140 & 0.002 & 0.003 \\
      & Maximum $E_u$ & 0.143 & 0.085 & 0.111 & 0.185 & 0.003 & 0.087 \\
    \bottomrule
  \end{tabular*}
  }
  \caption{Ablation of the adapter objective under the final benchmark
  configurations. The variants use identical architectures, observations,
  optimization budgets, and SVWS settings; the second sets
  $\lambda_\Phi=\lambda_t=0$ while retaining AdamW weight decay. Exact
  recovery and $E_u$ are computed over the same five seeds.}
  \label{tab:adapter_objective_ablation}
\end{table*}

Both objectives recover every S20 support and all KdV-T20 supports. Removing
the explicit penalties causes one KS-T20 failure and increases the maximum
field error from 0.037 to 0.085 on KdV-T20, from 0.108 to 0.185 on KS-T20,
and from 0.004 to 0.087 on 2D AD-T20, while the S20 errors remain comparable.
The explicit penalties therefore improve robustness primarily when temporal
coverage is sparse.

\FloatBarrier
\section{Support Dynamics under Joint Optimization}

To complement the representative trajectories in the main paper, we examine
all five test seeds for the three settings visualized there.
Table~\ref{tab:joint_mechanism_summary} also includes DeepMoD on KdV-S20, a
second setting in which the exact support persists to termination. The table
summarizes whether the exact support is reached at any recorded checkpoint, is
present at the checkpoint minimizing DeepMoD's fixed held-out objective, and
remains at termination. Figure~\ref{fig:joint_mechanism_all_seeds} shows the
complete support $F_1$ trajectories for the three settings in the main paper.

Because the two methods optimize different training objectives, we compare
their paths using support $F_1$ at each recorded checkpoint. DeepMoD support is
given by its learned sparsity mask at threshold $0.1$, whereas
Weak-PDE-LEARN (WPL) support is defined by coefficients whose absolute value
exceeds $10^{-3}$. For DeepMoD, the fixed held-out objective is the sum of data
MSE and PDE-residual MSE on its held-out split. Exact-support intervals are
derived from the resulting binary masks, and the recorded trajectories are
plotted without smoothing.

\begin{table}[t]
\centering
{\small
\setlength{\tabcolsep}{2.4pt}
\begin{tabular}{@{}llccc@{}}
\toprule
Method & Setting & Ever & Held-out min. & Terminal \\
\midrule
DeepMoD & KdV-S20 & 5/5 & 5/5 & 5/5 \\
DeepMoD & 2D AD-S20 & 5/5 & 5/5 & 5/5 \\
DeepMoD & KS-S100 & 0/5 & 0/5 & 0/5 \\
WPL & 2D AD-S20 & 5/5 & -- & 0/5 \\
\bottomrule
\end{tabular}
}
\caption{Exact support along recorded optimization paths, reported as runs out of five. Ever indicates an exact-support checkpoint; Held-out min. evaluates the DeepMoD checkpoint minimizing data MSE plus PDE-residual MSE on a fixed held-out split; Terminal denotes the final state. The held-out minimum is unavailable for Weak-PDE-LEARN because its test weak functions are resampled at each epoch. Ever is a retrospective path summary rather than a checkpoint selection rule.}
\label{tab:joint_mechanism_summary}
\end{table}

\begin{figure*}[t]
  \centering
  \includegraphics[width=\textwidth]{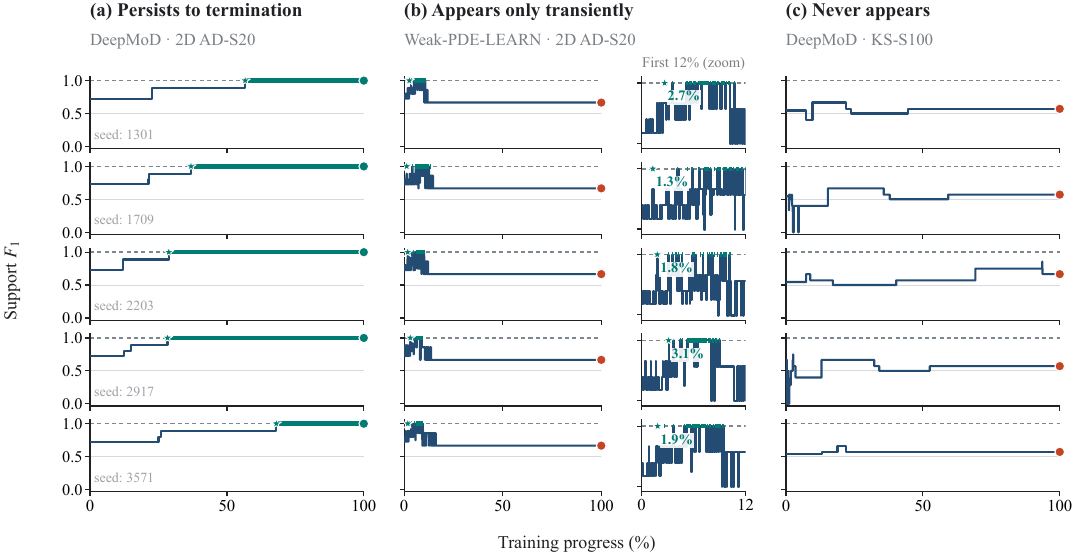}
  \caption{Support $F_1$ trajectories across all five seeds for the three
  settings shown in the main paper. Each row corresponds to one seed. The
  narrow subcolumn magnifies the first 12\% of Weak-PDE-LEARN training, with
  labels marking the first exact-support visit for each seed. Curves show
  unsmoothed support $F_1$ over training progress normalized within each
  recorded path. Teal segments mark exact-support intervals, stars mark the
  first exact visit, and terminal circles are teal when exact and vermilion
  otherwise.}
  \label{fig:joint_mechanism_all_seeds}
\end{figure*}

Across all five seeds, the trajectories exhibit the three behaviors
highlighted in the main paper. DeepMoD reaches and retains the exact support in
every KdV-S20 and 2D AD-S20 run. On KS-S100, it never reaches the exact support,
despite complete spatial coverage. Weak-PDE-LEARN reaches the exact support in
every 2D AD-S20 run but loses it before termination. Changing the checkpoint
can therefore alter the reported equation when the exact support is transient,
whereas checkpoint selection cannot recover a support that is absent
throughout the recorded path. These paths motivate separating field fitting
from equation selection: selection acts on one frozen reconstruction rather
than inheriting the support at a particular joint-training checkpoint.

In the representative Weak-PDE-LEARN trajectory shown in the main paper, the
exact support first appears at epoch 92 with
$\hat u_t=0.16u_x+0.52u_y+0.37u_{xx}+0.49u_{yy}$; at epoch 5000, the terminal
iterate retains only $\hat u_t=0.64u_y+0.77u_{yy}$.

\FloatBarrier

\section{Symbolic Discovery Beyond a Fixed Library}

The symbolic experiment tests whether the frozen reconstruction can also
support discovery when candidate expressions are generated rather than drawn
from a fixed library. The main paper reports aggregate recovery rates and
parameter estimates. Here we describe the benchmark and symbolic search
protocol, report example expressions from unsuccessful PySR runs, and list the
principal configuration settings. We fix the outer operator $\partial_{xx}$ and search for
the nonlinear diffusion function $q(u)$ in
\begin{equation}
  u_t=\partial_{xx}q^\star(u),
  \qquad q^\star(u)=0.1u^{1.73}
\end{equation}
Because $\partial_{xx}$ annihilates constants, $q(u)$ is identifiable only up
to an additive constant. We solve the equation on a periodic
$192\times121$ grid over $x\in[0,1)$ and $t\in[0,0.3]$.
The initial condition is
$u(x,0)=0.75+0.30\sin(2\pi x)+0.14\cos(4\pi x-0.35)
+0.07\sin(6\pi x+0.60)$.
We generate the trajectory using second-order periodic finite differences and
a BDF integrator with relative and absolute tolerances $2\times10^{-7}$ and
$10^{-9}$, respectively. At each noise level, the structured adapter is
trained on observations at 20\% fixed spatial sensors over all time points,
and five independent runs are evaluated. Independent zero-mean Gaussian noise
is added with standard deviation equal to 0\%, 5\%, or 10\% of the global
standard deviation of the clean field. The target exponent 1.73 is excluded
from the grid used to initialize the $u^p$ terminals.

Genetic programming (GP) constructs candidate expressions for $q(u)$. We run
two GP proposal searches, each using weak systems constructed from
independently shifted patch grids. Within each search, expressions with the
same canonical structure are grouped, and the expression with the smallest
normalized weak-form residual in each family is retained. The representatives
from the two searches are then merged and deduplicated. A separate weak system
fits the continuous parameters of each retained expression, and three
held-out weak systems evaluate the fitted candidates. SVWS selects from this
fixed candidate pool by minimizing the mean held-out normalized residual plus
$10^{-3}$ times expression complexity. The selected expression retains the
parameters fitted before validation.

A run counts as recovering the power-law family when the selected expression
is algebraically equivalent to
$\hat q(u)=\hat\kappa u^{\hat m}+C$ for some additive constant $C$. The
reported $\hat\kappa$ and $\hat m$ are summarized over runs that recover this
family. SVWS operates on the field reconstructed from 20\% fixed sensors. For
PySR, we first estimate $q(u)$ from the full noisy grid using weak equations
and a fixed basis of seven centered cubic B-splines in $u$, then apply PySR to
the resulting 512 $(u,\hat q(u))$ pairs. We use PySR 1.5.10 with
\emph{model\_selection=best}; the principal settings are listed in
Tables~\ref{tab:openform_svws_settings}
and~\ref{tab:openform_search_settings}.

Table~\ref{tab:openform_pysr_failures} shows expressions selected in
unsuccessful PySR runs, including forms with additional terms and polynomial
alternatives. These examples complement the aggregate recovery rates in the
main paper by illustrating the structures selected when the target power-law
family is not recovered.

\begin{table}[!t]
  \centering
  \small
  \begin{tabular*}{\columnwidth}{@{\extracolsep{\fill}}cc@{}}
    \toprule
    Noise (\%) & Example PySR expression \\
    \midrule
    0 & $0.100u^{1.727}+0.0002u^{2.727}-0.0001u+C$ \\
    5 & $0.0667u^2+0.0387u+C$ \\
    10 & $0.0624u^2+0.0437u+C$ \\
    \bottomrule
  \end{tabular*}
  \caption{Example PySR expressions from the unsuccessful run with the smallest prespecified seed at each noise level.}
  \label{tab:openform_pysr_failures}
\end{table}

\begin{table}[t]
  \centering
  \scriptsize
  \setlength{\tabcolsep}{2.5pt}
  \begin{tabularx}{\columnwidth}{@{}P{0.25\columnwidth}Y@{}}
    \toprule
    Component & Configuration \\
    \midrule
    Structured field adapter & 14 spatial features; 80 Fourier features; three
    128-unit hidden layers; 40 internal knots; Softplus output \\
    Adapter training & AdamW with weight decay $10^{-2}$; initial learning rate
    $10^{-3}$ with cosine annealing; gradient norm clipped to 1.0; at most 3200
    epochs; patience 600; at most 32 time frames per minibatch \\
    Adapter schedule & background warm start for 300 epochs; penalties enabled
    at epoch 500 and ramped over 900 epochs; terminal iterate \\
    Regularization & $(\lambda_\Phi,\lambda_t)=(3\times10^{-3},5\times10^{-4})$;
    weighted penalties capped at 0.12 and 0.20 times the observation loss \\
    GP primitives & leaves $u$, free constants, and $u^p$,
    $p\in\{0.5,0.6,\ldots,3.0\}$; operators
    $+$, $-$, $\times$, and protected division with sampling probabilities
    $(0.28,0.18,0.38,0.16)$ \\
    Protected operations & denominator floor $10^{-4}$; intermediate outputs
    clipped to $[-50,50]$; power input floor $10^{-6}$ \\
    \bottomrule
  \end{tabularx}
  \caption{Structured reconstruction and GP proposal settings for the
  symbolic experiment.}
  \label{tab:openform_svws_settings}
\end{table}

\begin{table}[t]
  \centering
  \scriptsize
  \setlength{\tabcolsep}{2.5pt}
  \begin{tabularx}{\columnwidth}{@{}P{0.25\columnwidth}Y@{}}
    \toprule
    Component & Configuration \\
    \midrule
    \multicolumn{2}{@{}l}{\textit{SVWS extension}} \\
    Proposal search & 2 independent GP searches; per search, separate
    128-patch fit and generation systems and one 192-patch proposal-validation
    system; $(h_x,h_t)=(0.105,0.03)$; kernel power $p=8$; population 72;
    16 generations \\
    Evolution & elite 8; mutation 0.35; crossover 0.55; maximum depth 6;
    complexity cap 7; complexity penalty $10^{-5}$ \\
    Parameter fit & 1 weak system; 128 patches;
    $(h_x,h_t)=(0.14,0.045)$; kernel power $p=8$ \\
    Continuous fit & at most 32 candidates; exponents $[0.2,4]$;
    constants $[-5,5]$; scale $[-10,10]$; at most six parameters, one
    division, and 35 function evaluations \\
    Validation & 3 systems; 192 patches each;
    $(h_x,h_t)=(0.105,0.03)$; kernel power $p=8$; 384 quadrature points;
    mean held-out risk plus $10^{-3}$ times expression complexity \\
    \addlinespace[2pt]
    \multicolumn{2}{@{}l}{\textit{PySR baseline}} \\
    PySR target construction & seven centered cubic B-spline basis functions
    in $u$; 512 weak patches; $(h_x,h_t)=(0.2,0.03)$; kernel power $p=6$;
    384 quadrature points; 512 state values; penalties of $10^{-6}$ for second
    differences and the additive offset \\
    PySR search & version 1.5.10; 100 iterations; 10 populations of 40;
    operators $(+,\times,\operatorname{power})$; maximum size 15 and depth 6;
    power exponent complexity at most 1; parsimony $10^{-4}$ \\
    \bottomrule
  \end{tabularx}
  \caption{Symbolic search, selection, and PySR settings.}
  \label{tab:openform_search_settings}
\end{table}

\FloatBarrier

\section{Sensitivity to Observation Density and Noise}
\label{sec:observation_sensitivity}

\subsection{Observation Density}

Figure~\ref{fig:observation_sweep} summarizes the observation-density study
for five methods spanning the proposed, classical, and neural approaches. KdV
is evaluated with 5\%, 10\%, 15\%, and 20\% fixed spatial sensors, whereas KS
and 2D AD are evaluated with 20\%, 50\%, 80\%, and 100\%. For each benchmark
and seed, the sensor masks are nested across densities. The same five test
seeds, method-specific configurations, and benchmark-specific candidate
libraries defined in Sec.~\ref{sec:supp_benchmarks} are used throughout each
sweep. The 20\% points are the runs reported in the main comparison. Curves
show medians and interquartile ranges over the five seeds.

\begin{figure*}[!t]
\centering
\includegraphics[width=0.88\textwidth]{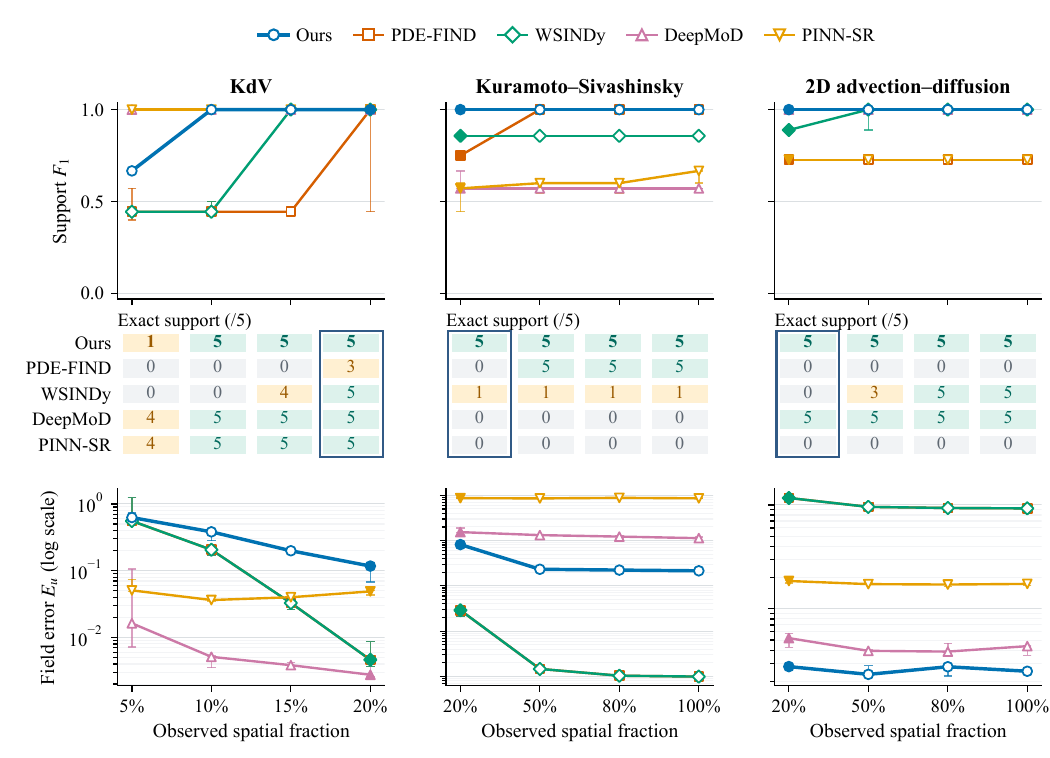}
\caption{Effect of observation density under clean, nested fixed-sensor sampling. Curves and error bars show the median and interquartile range of support $F_1$ and field error over five seeds. The middle panels report exact-support recoveries out of five. Filled markers and outlined columns identify the 20\% setting used in the main comparison.}
\label{fig:observation_sweep}
\end{figure*}

Increasing sensor density generally reduces field error, but the corresponding
gains in exact recovery depend on the method and PDE. Our method moves from 1/5 exact
recoveries at 5\% KdV sensors to 5/5 from 10\% onward, and remains at 5/5
throughout the KS and 2D AD sweeps. PDE-FIND on KS reaches 5/5 from 50\%
coverage, while WSINDy on 2D AD reaches 3/5 at 50\% and 5/5 from 80\%.
DeepMoD and PINN-SR do not recover the exact KS support at any tested density.
Thus, denser observations improve field reconstruction but do not by
themselves guarantee recovery of the correct equation terms.

\subsection{Observation Noise}

For the structured adapter with SVWS, we evaluate 10\% observation noise on KS
and 2D AD under both S20 and T20. Independent zero-mean Gaussian noise is added
to the sampled values with standard deviation equal to 10\% of the global
standard deviation of the clean field. The sampled coordinates and all adapter
and selector settings are identical to those in the corresponding clean runs.
Table~\ref{tab:supp_noise_robustness} compares the noisy and clean results.

\FloatBarrier

\noindent\begin{minipage}{\columnwidth}
\begin{center}
\centering
\footnotesize
\setlength{\tabcolsep}{1.5pt}
\begin{tabular}{llcccccc}
\toprule
& & \multicolumn{3}{c}{Clean} & \multicolumn{3}{c}{10\% noise} \\
\cmidrule(lr){3-5}\cmidrule(lr){6-8}
PDE & Obs. & Exact & $E_\xi$ & $E_u$ & Exact & $E_\xi$ & $E_u$ \\
\midrule
KS & S20 & 5/5 & 0.035 & 0.082 & 5/5 & 0.115 & 0.166 \\
KS & T20 & 5/5 & 0.055 & 0.082 & 5/5 & 0.083 & 0.097 \\
\midrule
2D AD & S20 & 5/5 & 0.010 & 0.003 & 5/5 & 0.011 & 0.006 \\
2D AD & T20 & 5/5 & 0.009 & 0.003 & 5/5 & 0.012 & 0.007 \\
\bottomrule
\end{tabular}
\captionof{table}{Robustness to 10\% observation noise. Exact reports recoveries out of five; errors are medians over the same seeds.}
\label{tab:supp_noise_robustness}
\end{center}

\end{minipage}

The structured adapter with SVWS recovers the exact support in all 20 noisy
runs. The largest increases in coefficient and field error occur on KS,
whereas the 2D AD estimates remain close to their clean counterparts. Thus, at
the tested noise level, support recovery is unchanged even though coefficient
and field accuracy deteriorate.

\FloatBarrier
\section{Code, Data, and Computational Resources}

Code and configurations will be made publicly available upon acceptance. The
release will provide the implementations, configurations, and run-level
results for the reported studies. It will include the structured
adapter and SVWS, baseline wrappers, observation records for the primary
fixed-library experiments, and scripts that verify the reported table values
and rebuild the numerical figures for optimization paths and observation
density. A README will provide execution commands, while a paper map will link
reported results to their implementations, manifests, and included evidence.
Environment files will list the required third-party dependencies.

The release will specify the public MDBench trajectory and state index and
include the script used to generate the symbolic diffusion benchmark.
Experiments were run on two Ubuntu servers. One server had an Intel Xeon Silver
4309Y CPU, 125~GiB of memory, and two 24-GiB NVIDIA RTX 4090 GPUs; the other had
an Intel Xeon w5-3525 CPU, 125~GiB of memory, and three 24-GiB NVIDIA RTX 4090
GPUs. Both environments ran Python 3.10. The first used PyTorch 2.5.1 with CUDA
12.1, and the second used PyTorch 2.7.1. Environment specifications and
versions of the principal dependencies will accompany the release. Neural
models were trained on one GPU per run using 32-bit floating-point model
tensors; source trajectories were stored in 64-bit precision.

\endgroup

\end{document}